\documentclass[runningheads]{llncs}

\usepackage{booktabs} 
\usepackage{lineno}
\usepackage{amsmath}
\usepackage{caption}
\usepackage{booktabs}
\usepackage{amsfonts}
\usepackage{algorithm}
\usepackage[noend]{algpseudocode}
\usepackage{longtable}
\usepackage{multicol}
\usepackage{rotating}
\usepackage{placeins}
 \usepackage{multirow}
\usepackage{hyperref}

\usepackage{xspace}
\usepackage{graphicx}
\usepackage{paralist}
\usepackage{url}
\usepackage{matharticle}
\usepackage{subfig}
\usepackage{etoolbox}

\usepackage{xcolor}
\usepackage{colortbl}
\definecolor{GAINSBORO}{HTML}{DCDCDC}
\definecolor{LIGHTYELLOW}{HTML}{FFFFE0}
\definecolor{MISTYROSE}{HTML}{FFE4E1}
\definecolor{ALICEBLUE}{HTML}{F0F8FF}

\algblock{ParFor}{EndParFor}
\algnewcommand\algorithmicparfor{\textbf{parfor}}
\algnewcommand\algorithmicpardo{\textbf{do}}
\algnewcommand\algorithmicendparfor{\textbf{end\ parfor}}
\algrenewtext{ParFor}[1]{\algorithmicparfor\ #1\ \algorithmicpardo}
\algrenewtext{EndParFor}{\algorithmicendparfor}

\algrenewcommand\alglinenumber[1]{\tiny #1:}

\newcommand{\horse}{\texttt{Lipizzaner}\xspace}

\newcommand{\Lipi}{\horse}

\newcommand{\spagan}{\texttt{SPaGAN}\xspace}
\newcommand{\nocommgan}{\texttt{IsoCoGAN}\xspace}
\newcommand{\pagan}{\texttt{PaGAN}\xspace}

\newcommand{\oppagan}{\texttt{PaGAN}\xspace}

\renewcommand{\v}[1]{\mathbf{#1}}

\newcommand{\xGrid}[1]{#1$\times$#1}

\newcommand{\nameAlgMixture}{\texttt{mixtureEvolution}\xspace}
\newcommand{\alg}{\texttt{Algorithm}\xspace}

\begin{document}
\title{Analyzing the Components of Distributed Coevolutionary GAN Training}

%
%
\author{Jamal Toutouh \and Erik Hemberg \and Una-May O'Reilly}
\authorrunning{J. Toutouh et al.}
%
\institute{Massachusetts Institute of Technology, CSAIL, MA, USA \\
\email{toutouh@mit.edu}, \email{\{hembergerik,unamay\}@csail.mit.edu}}
\maketitle              
\begin{abstract}
Distributed coevolutionary Generative Adversarial Network (GAN) training has empirically shown success in overcoming GAN training pathologies. This is mainly due to diversity maintenance in the populations of generators and discriminators during the training process. The method studied here coevolves sub-populations on each cell of a spatial grid organized into overlapping Moore neighborhoods. We investigate the impact on performance of two algorithm components that influence the diversity during coevolution: the performance-based selection/replacement inside each sub-population and the communication through migration of solutions (networks) among overlapping neighborhoods. In experiments on MNIST dataset, we find that the combination of these two components provides the best generative models. In addition, migrating solutions without applying selection in the sub-populations achieves competitive results, while selection without communication between cells reduces performance.

\end{abstract}
\section{Introduction}
\label{sec:introduction}

Machine learning with Generative Adversarial Networks (GANs) is a powerful method for generative modeling~\cite{goodfellow2014generative}. 
A GAN consists of two neural networks, a generator and a discriminator, and applies adversarial learning to optimize their parameters. 
The generator is trained to transform its inputs from a random latent space into ``artificial/fake'' samples that approximate the true distribution. 
The discriminator is trained to correctly distinguish the ``natural/real'' samples from the ones produced by the generator.  
Formulated as a minmax optimization problem through the definitions of generator and discriminator loss, training can converge on an optimal generator that is able to fool the discriminator.

GANs are difficult to train. The adversarial dynamics introduce convergence pathologies~\cite{arora2017generalization,li2017towards}.
This is mainly because the generator and the discriminator are differentiable networks, 
their weights are updated by using (variants of) simultaneous gradient-based methods to optimize the minmax objective, that rarely converges to an equilibrium. 
Thus, different approaches have been proposed to improve the convergence and the robustness in GAN training~\cite{ganhacks,mao2017least,nguyen2017dual,radford2015unsupervised,toutouh2019}.

\sloppy
A promising research line is the application of distributed competitive coevolutionary algorithms (Comp-COEA). 
Fostering an arm-race of a population of generators against a population of discriminators, these methods optimize the minmax objective of GAN training. 
Spatially distributed populations (cellular algorithms) are effective at mitigating and resolving the COEAs pathologies attributed to a lack of diversity~\cite{popovici2012coevolutionary}, which are similar to the ones observed in GAN training.
\Lipi~\cite{schmiedlechner2018towards,schmiedlechner2018lipizzaner} is a spatial distributed Comp-COEA that locates the individuals of both populations on a spatial grid (each cell contains a GAN). 
In a cell, each generator is evaluated against all the discriminators of its neighborhood, the same with the discriminator. 
It uses neighborhood communication to propagate models and foster diversity in the sub-populations. 
Moreover, the selection pressure helps the convergence in the sub-populations~\cite{alba2009cellular}. 

Here, we evaluate the impact of neighborhood communication and selection pressure on this type of GAN training. 
We conduct an ablation analysis to evaluate different combinations of these two components. 
We ask the following research questions: 
\textbf{RQ1:} \textit{What is the effect on the quality of the generators when training with communication or isolation and the presence or absence of selection pressure?}. The quality of the generators is evaluated in terms of the accuracy of the samples generated and their diversity. 
\textbf{RQ2:} \textit{What is the effect on the diversity of the network parameters when training with communication or isolation and the presence or absence of selection pressure?} 
\textbf{RQ3:} \textit{What is the impact on the computational cost of applying migration and selection/replacement?}

The main contributions of this paper are: 
\textit{i)} proposing distributed Comp-COEA GAN training methods by applying different types of ablation to \Lipi, \textit{ii)} evaluating the impact of the communication and the selection pressure on the quality of the returned generative model, and \textit{iii)} analyzing their computational cost.

The paper is organized as follows. Section~\ref{sec:related-work}
presents related work. 
Section~\ref{sec:methods} describes \Lipi and the ablations the methods analyzed.
The experimental setup is in
Section~\ref{sec:experimental-setup} and results in
Section~\ref{sec:results}. Finally, conclusions are drawn and future
work is outlined in Section~\ref{sec:conclusions}.

\vspace{-0.4cm}
\section{Related Work}
\label{sec:related-work}
\vspace{-0.4cm}

In 2014, Goodfellow introduced GAN
training~\cite{goodfellow2014generative}. Robust GAN training methods
are still investigated~\cite{arora2017generalization,li2017towards}.
Competitive results have been provided by several practices that stabilize the training~\cite{ganhacks}. These methods include different strategies, such as
using different cost functions to the generator or
discriminator~\cite{arjovsky2017wasserstein,mao2017least,nguyen2017dual,zhao2016energy,wang2019evolutionary} and decreasing the learning rate through the iterations~\cite{radford2015unsupervised}. 

GAN training that involves multiple
generators and/or discriminators empirically show robustness. 
Some examples are:
iteratively training and adding new generators with boosting
techniques~\cite{tolstikhin2017adagan}; combining a cascade of
GANs~\cite{wang2016ensembles}; training an array of discriminators on different low-dimensional projections of the
data~\cite{neyshabur2017stabilizing}; training the generator against a
dynamic ensemble of discriminators~\cite{mordido2018dropout};
independently optimizing several ``local'' generator-discriminator
pairs so that a ``global'' supervising pair of networks can be trained
against them~\cite{chavdarova2018sgan}. 

Evolutionary algorithms (EAs) may show limited effectiveness in high dimensional problems because of runtime~\cite{talbi2009metaheuristics}. 
Parallel/distributed implementations allow EAs to keep computation times at reasonable levels~\cite{alba2013parallel,essaid2019gpuea}.
There has been an emergence of large-scale distributed evolutionary machine learning systems. For example: EC-Star~\cite{hodjat2014maintenance}, which
runs on hundreds of desktop machines; a simplified version of Natural
Evolution Strategies~\cite{wierstra2008natural} with a novel
communication strategy to address a collection of reinforcement learning benchmark problems~\cite{salimans2017evolution} or deep convolutional networks trained with genetic algorithms~\cite{clune2017uber}.  

Theoretical studies and empirical results demonstrate that
the spatially distributed COEA GAN training mitigates convergence pathologies~\cite{schmiedlechner2018towards,schmiedlechner2018lipizzaner,toutouh2019}. 
In this study, we focus on spatially distributed Comp-COEA GAN
training, such as \Lipi~\cite{schmiedlechner2018towards,schmiedlechner2018lipizzaner}. 
\Lipi places the individuals of the generator and discriminator populations on each cell (i.e., each cell contains a generator-discriminator pair). 
Overlapping Moore neighborhoods determine the communication among the cells to propagate the models through the grid. 
Each generator is evaluated against all the discriminators of its neighborhood and the same happens with each discriminator. 
This intentionally fosters diversity to address GAN training pathologies.
Mustangs~\cite{toutouh2019}, a \Lipi variant, uses randomly selected loss functions to train each cell for each epoch to increase diversity. 
Moreover, training the GANs in each cell with different subsets of the training dataset has been demonstrated effective in increasing diversity across the grid~\cite{toutouh2020_deepneuroevolution}.  
These three approaches return an ensemble/mixture of generators defined by the best neighborhood (sub-population of generators) built using evolutionary ensemble learning~\cite{Toutouh_GECO2020}.  
They have shown competitive results on standard benchmarks.


Here, we investigate the impact of two key components for
diversity in Comp-COEA GAN training using ablations. 

\vspace{-.3cm}
\section{Comp-COEA GAN Training Ablations}
\label{sec:methods}
\vspace{-0.2cm}

Here, we evaluate the impact of communication and selection
pressure on the performance of distributed Comp-COEA GAN
training. 
One goal of the training is to maintain diversity as a means of resilience to GAN training pathologies. 
The belief is that a lack of sufficient diversity results in convergence to pathological training states~\cite{schmiedlechner2018lipizzaner} and too much diversity results
in divergence. The use of selection and communication is
one way of regulating the diversity.

This section presents \Lipi~\cite{schmiedlechner2018lipizzaner} and describes the ablations applied to gauge the impact of communication/isolation and selection pressure.

\vspace{-0.3cm}
\subsection{Lipizzaner Training}
\label{sec:lipizaner}
\vspace{-0.2cm}

\begin{figure}[ht!]
\vspace{0.0cm}
\centering
  \includegraphics[width=0.6\linewidth]{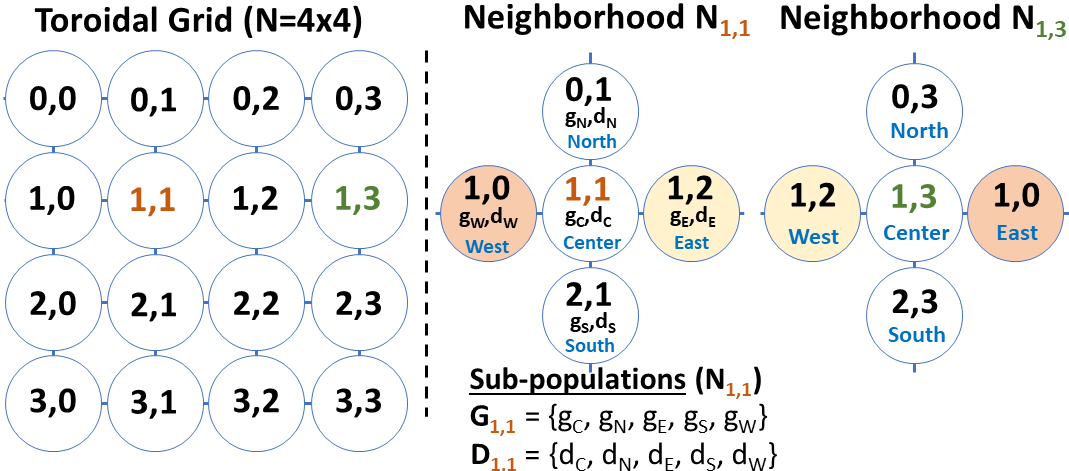}
  \vspace{-0.1cm}
  \caption{A 4$\times$4 grid (left) and the neighborhoods N$_{1,1}$ and N$_{1,3}$ (right).}
  \label{fig:4x4-nhood}
\vspace{0.cm}
\end{figure}

\Lipi adversarially trains a population of generators $\mathbf{g}=\{g_1, ..., g_N\}$ and a population of discriminators $\mathbf{d}=\{d_1, ..., d_N\}$, where $N$ is the size of the populations. 
It defines a toroidal grid in whose cells a pair generator-discriminator, i.e., a GAN, is placed (called \textit{center}).
This allows the definition of neighborhoods with sub-populations $\mathbf{g}^{k}$ and $\mathbf{d}^{k}$, of $\mathbf{g}$ and of $\mathbf{d}$,  respectively. 
The size of these sub-populations is denoted by $s$ ($s \leq N$). 
\Lipi uses the five-cell Moore neighborhood ($s=5$), i.e., the neighborhoods include
the cell itself (\textit{center}) and the cells in the \textit{West},
\textit{North}, \textit{East}, and \textit{South} (see
Figure~\ref{fig:4x4-nhood}).

Each cell asynchronously executes in parallel its own learning algorithm. Cells interact with the neighbors at the beginning of each training epoch.
This communication is carried out by gathering the
latest updated \textit{center} generator and discriminator of its
overlapping neighborhoods. 
Figure~\ref{fig:4x4-nhood}
illustrates some examples of the overlapping neighborhoods on a
\textit{4x4} toroidal grid. The updates in cell N$_{1,0}$ and N$_{1,2}$ will be communicated to the N$_{1,1}$ and N$_{1,3}$ neighborhood.\vspace{-5pt}

\Lipi returns an ensemble of generators that consist of a sub-population of generators $\mathbf{g}^{k}$ and a mixture weight vector $\v{w}\subset \R^{s}$, where $w_i \in \v{w}$ represents the probability that a data point is drawn from $\mathbf{g}^{k}_i$, i.e., $\sum_{w_i \in \v{w}^k} w_i = 1$.
Thus, an Evolutionary Strategy, \texttt{ES}-(1+1), evolves $\v{w}$ to optimize the weights in order to get the most accurate generative model according to a given metric~\cite{schmiedlechner2018lipizzaner}, such as Fr\'echet Inception Distance (FID)~\cite{heusel2017gans}.\vspace{-5pt} 

\alg~\ref{alg:general-coev} summarizes the main steps of \Lipi. 
It starts the parallel execution of the
training process in each cell.  First, the cell randomly initializes
single generator and discriminator models.  Then, the training process
consist of a loop with three main steps: \textit{i)}
gathering the GANs (neighbors) to update the neighborhood; 
\textit{ii)} updating the \textit{center} by applying the training method presented in 
\alg~\ref{alg:lipi};
and \textit{iii)} evolving mixture of weights by applying \texttt{ES}-(1+1) (line~\ref{alg:coev-weights} in \alg~\ref{alg:general-coev}). 
These three steps are repeated $T$ (training epochs) times.  
The returned generative model is the best performing neighborhood with its optimal mixture weights, i.e., the best ensemble of generators.\vspace{-5pt} 
 
The Comp-COEA training applied to the networks of each sub-population is shown in \alg~\ref{alg:lipi}. 
The output is the sub-population $n$ with the \textit{center} updated. 
This method is built on the basis of \textit{fitness evaluation}, \textit{selection and replacement}, and \textit{reproduction} based on gradient descent updates of the weights.\vspace{-5pt}

The \textit{fitness} of each network is evaluated in terms of the average \textit{loss function} $\calL$ when it is evaluated against all the adversaries. 
After evaluating all the networks, a \textit{tournament selection operator} is applied to select the parents (a generator and a discriminator) to generate the offspring (lines from \ref{alg:lipi-get-minibatch} to \ref{alg:lipi-select-center}). \vspace{-5pt}

The selected generator/discriminator is evaluated against a randomly chosen adversary from the sub-population (lines \ref{alg:lipi-rand-d} and \ref{alg:lipi-rand-g}, respectively). 
The computed losses are used to mutate the parents (i.e., update the networks' parameters) according to the stochastic gradient descent (SGD), which learning rate values $n_{\delta}$ were updated applying Gaussian-based mutation (line~\ref{alg:lipi-update-lr}).\vspace{-5pt}

When the training is completed, all models are evaluated
again, the least fit generator and discriminator
in the sub-populations are replaced with the fittest ones and sets them as the \textit{center} of the cell (lines from \ref{alg:fit-eval-1.5} to
\ref{alg:replace-center}). \vspace{-5pt}

\setlength{\textfloatsep}{0pt}
\begin{algorithm}
	\small
	\caption{\Lipi\newline
		\textbf{Input:}~T: Training epochs, ~$E$: Grid cells, ~$s$: Neighborhood size,
		~$\theta_{EA}$: \nameAlgMixture parameters, ~$\theta_{Train}$: Parameters for training the models in each cell\newline
		\textbf{Return:}
		~$g$: Neighborhood of generators, ~$\mathbf{w}$: Mixture weights
	}\label{alg:general-coev}
	\begin{algorithmic}[1]
		
		\ParFor{$c \in E$} \Comment{Asynchronous parallel execution of all cells in grid}
		\State $n, \mathbf{w} \gets$ initializeNeighborhoodAndMixtureWeights($c, s$) 
		\For{epoch} $\in [1,\dots, \text{T}]$ \Comment{Iterate over training epochs}
                \State $n \gets$ copyNeighbours($c, s$) \Comment{Gather neighbour networks} \label{alg:copy-neighboor}
		\State $n \gets$ trainModels($n, \theta_{Train}$)\Comment{Update GANs weights} \label{alg:coev-training}
		\State $g \gets$ {getNeighborhoodGenerators}($n$)\Comment{Get the generators} 
		\State $\mathbf{w} \gets$ \nameAlgMixture($\mathbf{w}, g, \theta_{EA}$)\Comment{Evolve mixture weights by \texttt{ES}-(1+1)}\label{alg:coev-weights}
		\EndFor
		\EndParFor
		\State $(g, \mathbf{w}) \gets$ \texttt{bestEnsemble}($g^*, \mathbf{w}^*$)\Comment{Get the best ensemble} 
		\State \Return $(g, \mathbf{w})$ \Comment{ Cell with best generator mixture } 
	\end{algorithmic}
\end{algorithm}

\setlength{\textfloatsep}{0pt}
\begin{algorithm}
	\small
	\caption{\texttt{Coevolve and train the networks} \\
		\textbf{Input:}
		~$\tau$: Tournament size, ~$X$: Input training dataset,
		~$\beta$: Mutation probability, ~$n$: Cell neighborhood sub-population, ~$M$: Loss functions \newline
		\textbf{Return:}
		~$n$: Cell neighborhood sub-population updated
	}\label{alg:lipi}
	\begin{algorithmic}[1]
	\State $\mathbf{B} \gets $ getMiniBatches($X$) \Comment{ Load minibatches } \label{alg:lipi-get-minibatch}
	\State $B \gets $ getRandomMiniBatch($\mathbf{B}$) \Comment{ Get a minibatch to evaluate GANs} \label{alg:get-ran-batch-1}	
	\For{$g, d \in \mathbf{g} \times \mathbf{d}$} \Comment{Evaluate all GAN pairs} \label{alg:for-selec-1}
		\State $\mathcal{L}_{g,d} \gets$ evaluate($g, d, B$) \Comment{ Evaluate GAN} \label{alg:fit-eval-1}
		\EndFor
	  \State $g, d \gets$ select($n, \tau$) \Comment{Select with min loss($\mathcal{L}$) as fitness } \label{alg:lipi-select-center}
		
		\For{$B \in \mathbf{B}$} \Comment{Loop over batches}
		\State $n_{\delta} \gets$ mutateLearningRate($n_{\delta}, \beta$) \Comment{ Update learning rate}\label{alg:lipi-update-lr}
		\State $d' \gets$ getRandomOpponent($\mathbf{d}$) \Comment{Get random discriminator to train $g$} \label{alg:lipi-rand-d}
		\State $\nabla_{g} \gets$ computeGradient($g, d'$) \Comment{ Compute gradient for $g$ against $d'$ }
		\State $g \gets$ updateNN($g, \nabla_g, B$) \Comment{ Update $g$ with gradient }
		\State $g' \gets$ getRandomOpponent($\mathbf{g}$) \Comment{ Get uniform random generator to train $g$} \label{alg:lipi-rand-g}
		\State $\nabla_{d} \gets$ computeGradient($d, g'$) \Comment{ Compute gradient for $d$ against $g'$}
		\State $d \gets$ updateNN($d, \nabla_d, B$) \Comment{ Update $d$ with gradient }
		\EndFor
		
		\For{$g, d \in \mathbf{g} \times \mathbf{d}$} \Comment{Evaluate all updated GAN pairs} \label{alg:fit-eval-1.5}
		\State $\mathcal{L}_{g,d} \gets$ evaluate($g, d, B$) \Comment{ Evaluate GAN} \label{alg:fit-eval-2}
		\EndFor
		\State $\mathcal{L}_{g} \gets average(\mathcal{L}_{\cdot, d})$ \Comment{ Fitness of $g$ is the average loss value ($\mathcal{L}$)}
		\State $\mathcal{L}_{d} \gets average(\mathcal{L}_{g,\cdot})$ \Comment{ Fitness of $d$ is the average loss value ($\mathcal{L}$)}    
		\State $n \gets$ replace($n, \mathbf{g}$) \Comment{ Replace the generator with worst loss}
		\State $n \gets$ replace($n, \mathbf{d}$) \Comment{ Replace the discriminator worst loss}
                \State $n \gets$ setCenterIndividuals($n$) \Comment{Best $g$ and $d$ are placed in the center} \label{alg:replace-center}
		\State \Return $n$
	\end{algorithmic}
\end{algorithm}

\vspace{-0.22cm}
\subsection{Lipizzaner Ablations Analyzed}
\label{sec:ablations}
\vspace{-0.1cm}

We conduct an ablation analysis of \Lipi to evaluate the impact of different degrees of communication/isolation and selection pressure on its COEA GAN training. The ablations are listed in Table~\ref{tab:ablations}. They ablate the use of \textit{sub-populations}, the \textit{communication} between cells, and the application of the \textit{selection/replacement} operator. 
Thus, we define three variations of \Lipi: 
\vspace{-5mm}
\begin{itemize}
\item Spatial Parallel GAN training (\spagan): It does not apply selection/replacement. After gathering the networks from the neighborhood, it uses them to only train the \textit{center} (i.e., \spagan does not apply the operations in lines between~\ref{alg:get-ran-batch-1} and~\ref{alg:lipi-select-center} and lines between~\ref{alg:fit-eval-1.5} and~\ref{alg:replace-center} of~\alg~\ref{alg:lipi}).

\item Isolated Coevolutionary GAN training (\nocommgan): It trains sub-populations of GANs without communication between cells. There is no exchange of networks between neighbors after the creation of the initial sub-population (i.e., the operation in line~\ref{alg:copy-neighboor} of \alg~\ref{alg:general-coev} is applied only during the first iteration of its main loop).

\item Parallel GAN (\pagan): This trains a population of $N$ GANs in parallel. 
When all the GAN training is finished, it randomly produces $N$ sub-sets of $s\leq N$ generators selected from the entire population of trained generators to define the ensembles and optimizes the mixture weights with ES-(1+1). 
\end{itemize}

\begin{table}[!h]
\vspace{-1.3cm}
\setlength{\tabcolsep}{3pt} 
\renewcommand{\arraystretch}{0.9} 
  \centering
  \scriptsize
  \caption{\small Key components of the GAN training methods.}
  \label{tab:ablations}
\begin{tabular}{lrrrr}
\toprule
\textbf{Feature} &  \textbf{\Lipi} &  \textbf{\spagan} &  \textbf{\nocommgan} &   \textbf{\pagan} \\
\midrule
\textbf{Use of sub-populations} & $\checkmark$  & $\checkmark$ & $\checkmark$ & -\\
\textbf{Communication between sub-populations}  & $\checkmark$ & $\checkmark$ &  -& -\\
\textbf{Application of selection/replacement operator}  & $\checkmark$ & - & $\checkmark$ & -\\
\bottomrule
\end{tabular}
  \vspace{-.7cm}
\end{table}

\vspace{-.4cm}
\section{Experimental Setup}
\label{sec:experimental-setup}
\vspace{-.3cm}
This experimental analysis compares \Lipi, \spagan, \nocommgan, and \pagan in creating generative models to produce samples of the MNIST dataset~\cite{lecun1998mnist}. 
This dataset is widely used as a benchmark due to its target space and dimensionality. 

The communication among the neighborhoods is affected by the grid
size~\cite{schmiedlechner2018lipizzaner}. We evaluate \spagan and
\Lipi by using three different grid sizes: \xGrid{3}, \xGrid{4}, and
\xGrid{5}, to control for the impact of this parameter.

To evaluate the quality of the generated data we use \textbf{FID} score. 
The \textbf{network diversity} of the trained models is evaluated by the \textbf{$L_2$ distance between the parameters} of the generators.  The \textbf{total variation distance (TVD)}~\cite{li2017distributional}, which is a scalar that measures class
balance, is used to analyze the diversity of the generated data by the generative models.
TVD reports the difference between the proportion of the generated
samples of a given digit and the ideal proportion (10\% in MNIST).
Finally, the \textbf{computational cost} is measured in terms of run time. All the analyzed methods apply the same number of training-network steps, i.e., updating the networks’ parameters according to SGD.    
All implementations are publicly available\footnote{\Lipi and ablations - \texttt{https://github.com/ALFA-group/lipizzaner-gan}} and use the same \texttt{Python} libraries and versions.
The distribution of the results is not Gaussian, so a Mann-Whitney U
statistical test is applied to evaluate their significance.

The methods evaluated here are configured according to the settings
proposed by the authors of \Lipi~\cite{schmiedlechner2018lipizzaner}. 
The experimental analysis is performed on a cloud computing platform that provides 8~Intel Xeon cores 2.2GHz with 32 GB RAM and an NVIDIA Tesla P100 GPU with 16 GB RAM.

\vspace{-0.3cm}
\section{Results and Discussion}
\label{sec:results}
\vspace{-0.3cm}


This section discusses the main results of the experiments carried out
to evaluate \Lipi, \spagan, \nocommgan, and \pagan.

\vspace{-0.2cm}
\subsection{Generator Quality}
\label{sec:final-fid}
\vspace{-0.2cm}
Table~\ref{tab:final-fids} summarizes the results by showing the best
FID scores for the different methods and grid sizes in the
30~independent runs. 
\Lipi obtains the lowest/best
\textbf{Mean}, \textbf{Median}, and \textbf{Min} FID scores for all
grid sizes.  \spagan and \pagan are the second and third best methods,
respectively.  \nocommgan presents the lowest quality by showing the
highest (worst) FID scores. 
The FID values in terms of hundreds indicate the generators are not capable of creating adequate MNIST data samples. 
The Mann-Whitney U test shows that the methods that exchange individuals among the neighborhoods, i.e.,
\Lipi and \spagan, are significantly better than \pagan and \nocommgan
($\alpha<<0.001$). These results are confirmed by posthoc statistical
analysis.  According to this analysis there is no statistical
difference between \Lipi \xGrid{3} and \spagan for all evaluated grid
sizes. In turn, \Lipi \xGrid{4} and \xGrid{5} outperform all
the other methods and \Lipi \xGrid{5} provides the best generators (lowest FID).  

\begin{table}[!h]
\setlength{\tabcolsep}{4pt}
\renewcommand{\arraystretch}{0.85}
	\centering
	\small
	\vspace{-1.cm}
	\caption{\small FID results (Low FID indicates more quality)}
	\label{tab:final-fids}
\begin{tabular}{llrrrrr}
\toprule
\textbf{Grid} & \textbf{Method} &  \textbf{Mean}$\pm$\textbf{Std} &  \textbf{Median} &   \textbf{Iqr} &   \textbf{Min} &   \textbf{Max} \\
\midrule
\multirow{4}{*}{\xGrid{3}} & \Lipi & \textbf{40.93$\pm$8.51} & \textbf{39.44} &  10.59 &  \textbf{28.04}  &  62.10 \\ 
&\spagan & 43.59$\pm$5.53 & 43.09 &  5.94 &  30.65  &  \textbf{51.81} \\ 
&\nocommgan & 881.79$\pm$52.67 & 871.04 &  66.81 &  798.03  &  998.79 \\ 
&\oppagan & 51.15$\pm$14.06 & 46.85 &  6.40 &  40.81  &  112.58 \\ 
\midrule
\multirow{2}{*}{\xGrid{4}} & 
\Lipi & \textbf{32.84$\pm$6.93} &\textbf{ 32.22} &  7.23 &  \textbf{19.15}  &  \textbf{46.53} \\
&\spagan & 37.97$\pm$8.89 & 35.98 &  13.61 &  23.69  &  56.36 \\ 
\midrule
\multirow{2}{*}{\xGrid{5}} &
\Lipi & \textbf{28.74$\pm$4.91} & \textbf{28.14} &  7.45 &  \textbf{22.56}  &  \textbf{40.57} \\
&\spagan  & 39.11$\pm$4.00 & 39.61 &  5.29 &  27.39  &  48.03 \\ 
\bottomrule
\end{tabular}
\vspace{-.75cm}
\end{table}

\nocommgan does not converge since the individuals of one
sub-population are evaluated against randomly chosen individuals of the other one. As the accuracy of their fitness evaluation depends subjectively on the quality of the randomly chosen opponent, it is likely that the fitness value does not
correspond to the real quality of the individual, and therefore, the
selection/replacement operator does not promote the objectively best solution in the sub-population. 

For all grid sizes, \spagan provides higher/worse FIDs than \Lipi. 
\spagan has a similar quality for all grid sizes, but \Lipi improves when the grid size is increased.  Thus, the
larger the grid size, the greater the difference between these two
methods.  We observe the benefits of increasing diversity
(population/grid size) when applying the coevolutionary approach with
selection and replacement of \Lipi. The results of \Lipi are mainly
due to, first, the larger population sizes' ability to encompass a
higher diversity, and second, the selection/replacement process
applied by \Lipi accelerates the convergence of the population to higher quality generators.


\vspace{-0.35cm}
\subsection{Quality (FID) Evolution}
\label{sec:fid-evolution}
\vspace{-0.1cm}

\begin{figure*}
\vspace{-1.7cm}
\centering
\begin{minipage}[!h]{0.43\textwidth}
\begin{figure}[H]
\centering
\includegraphics[width=\textwidth]{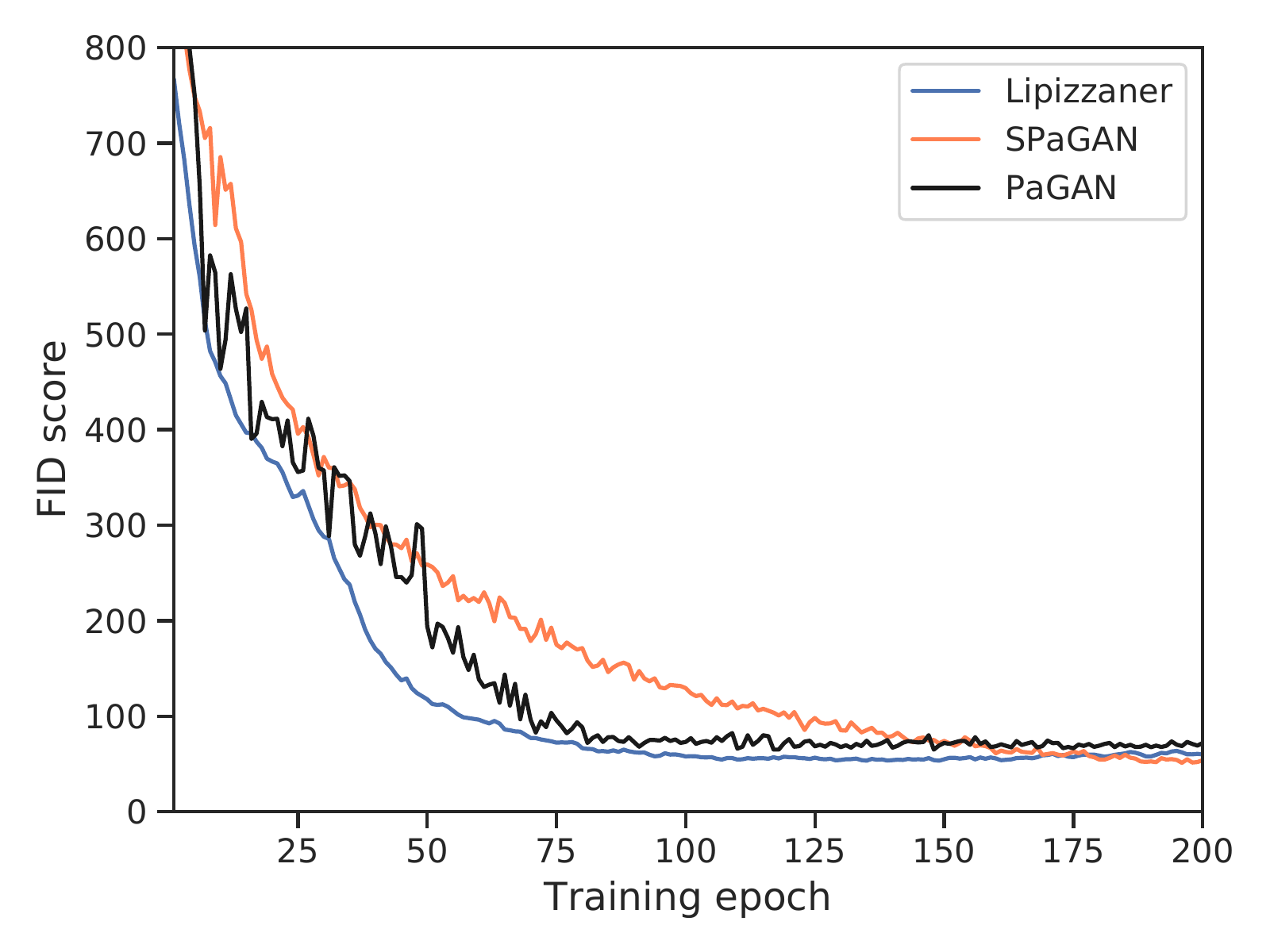}
\vspace{-0.7cm}
\caption{Median FID evolution through the 200 epochs (\xGrid{3}).}\label{fig:fid-gridsize-evolution}
\end{figure}
\end{minipage}
\hspace{0.08\textwidth}
\begin{minipage}[!h]{0.43\textwidth}
\begin{figure}[H]
\centering
\includegraphics[width=\textwidth]{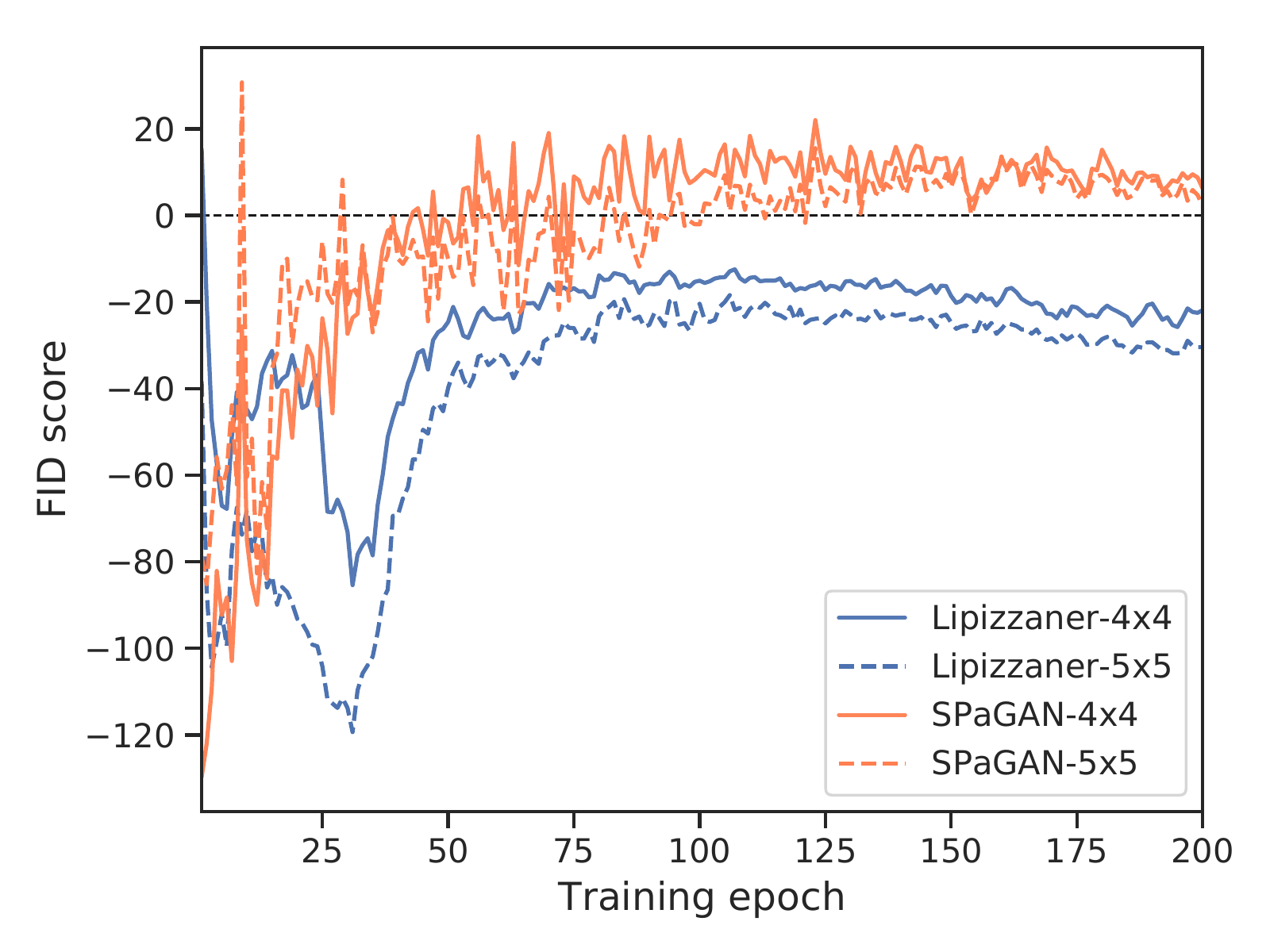}
\vspace{-0.7cm}
\caption{FID differences when using differnt grids ($diff\_FID^{m \times m}_{i}$).}\label{fig:fid-differences}
\end{figure}
\end{minipage}
\vspace{-.5cm}
\end{figure*}
\begin{figure*}[h!]
\vspace{-0.4cm}

\centering
\subfloat[\scriptsize{\spagan(25)}]
         {\includegraphics[width=0.2\textwidth]{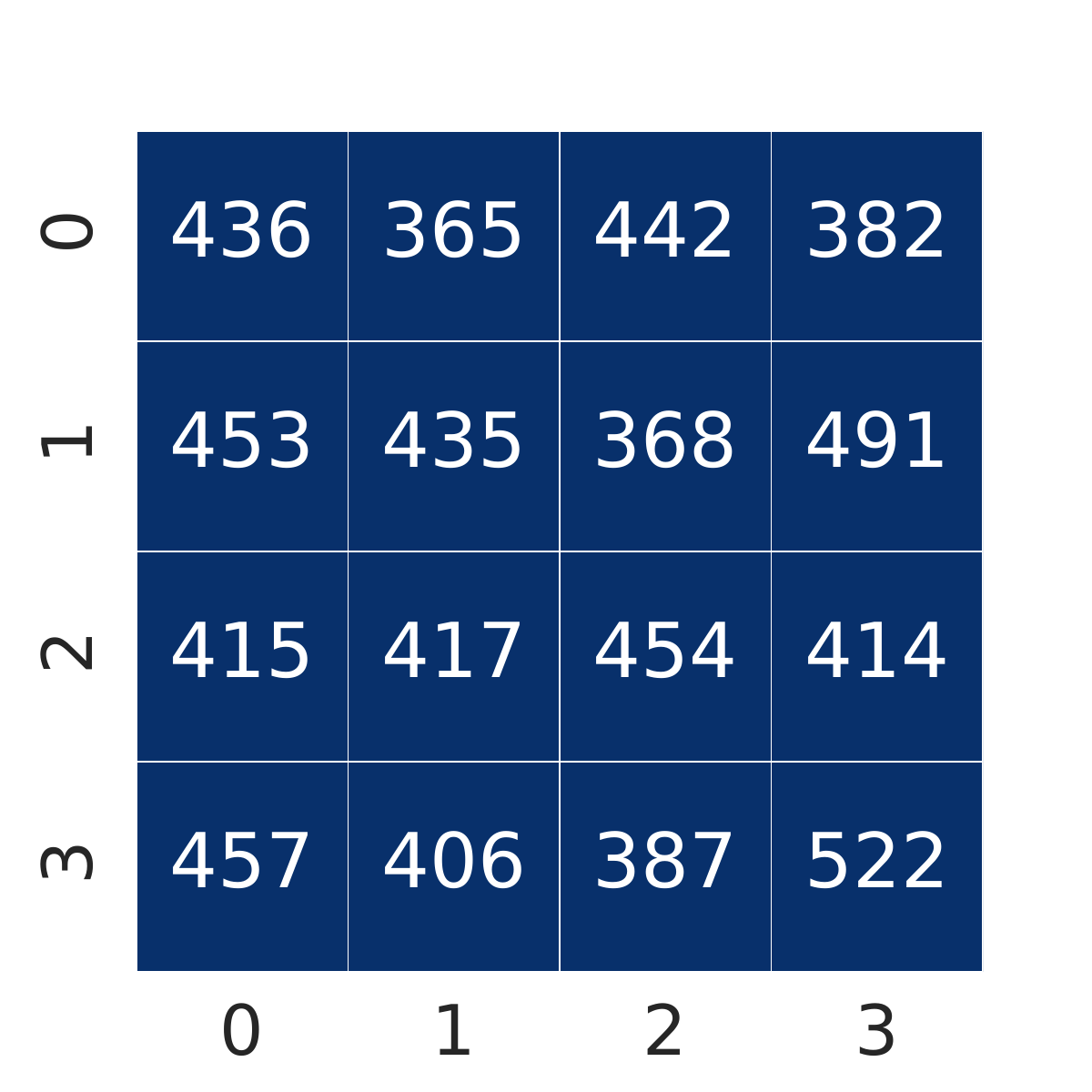}\label{fig:heatmpa-mnist-4x4-025-unif}}\hspace{0.1cm}
\subfloat[\scriptsize{\spagan(50)}]
         {\includegraphics[width=0.2\textwidth]{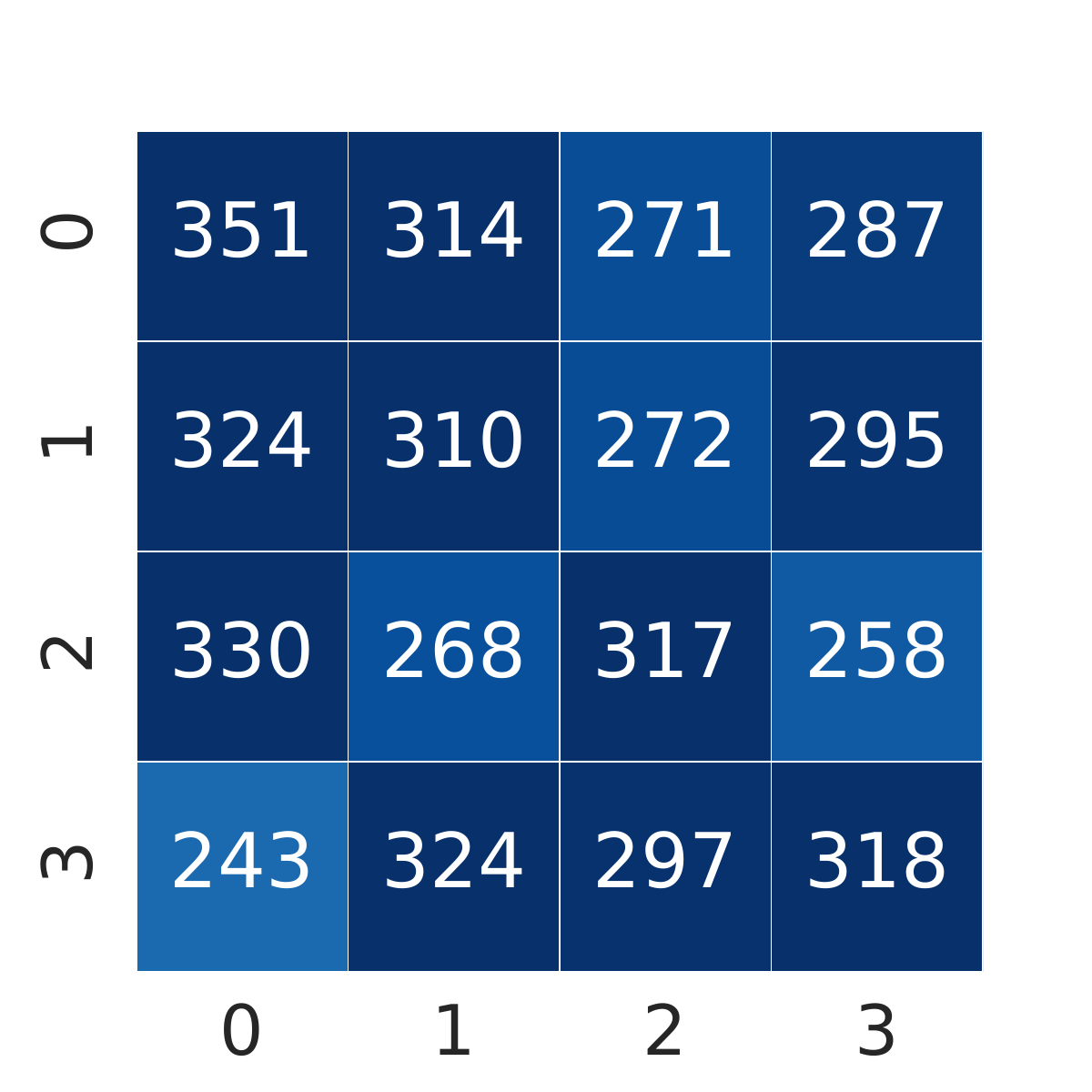}\label{fig:heatmpa-mnist-4x4-050-unif}}\hspace{0.1cm}
\subfloat[\scriptsize{\spagan(75)}]
         {\includegraphics[width=0.2\textwidth]{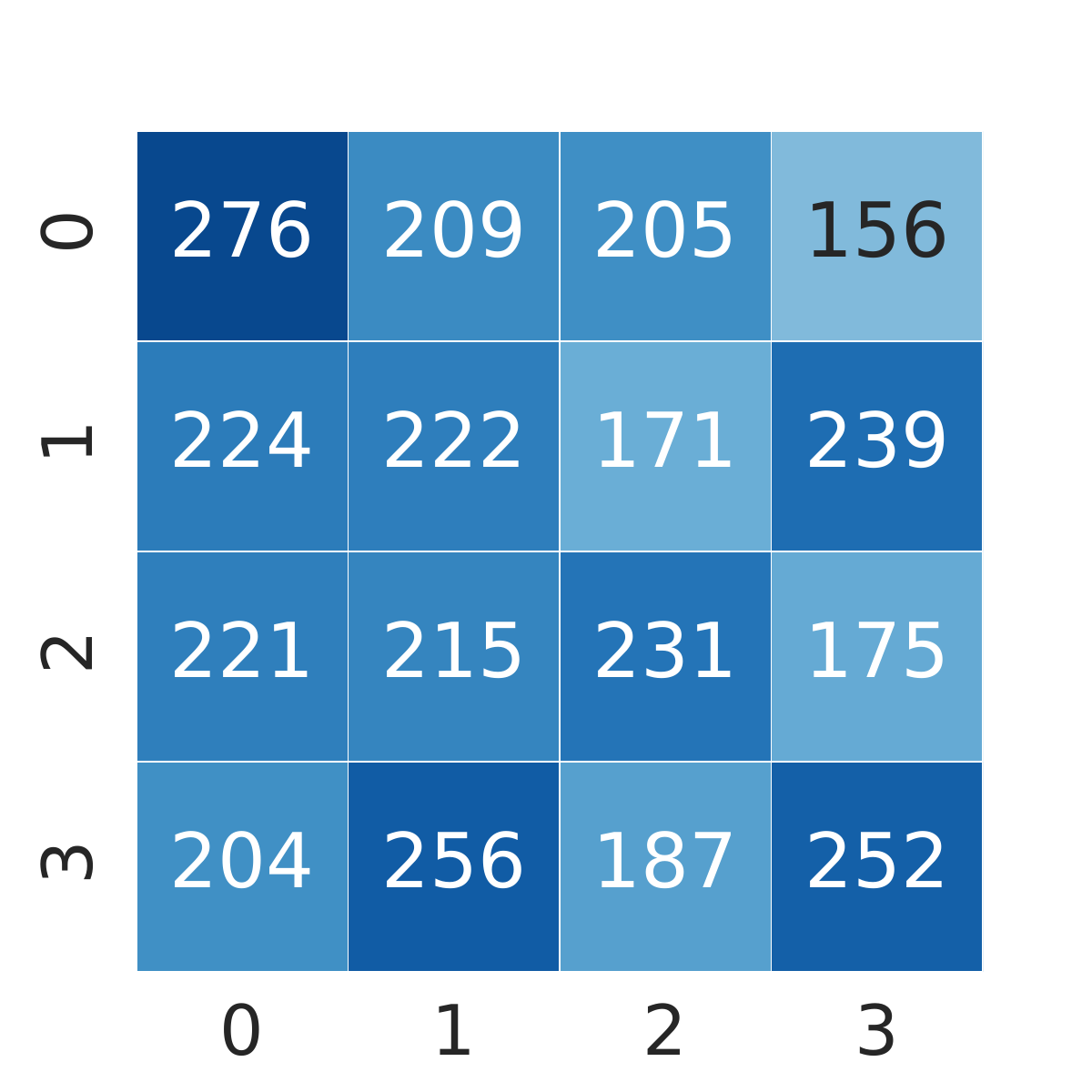}\label{fig:heatmpa-mnist-4x4-075-unif}}\hspace{0.1cm}
\subfloat[\scriptsize{\spagan(100)}]
         {\includegraphics[width=0.2\textwidth]{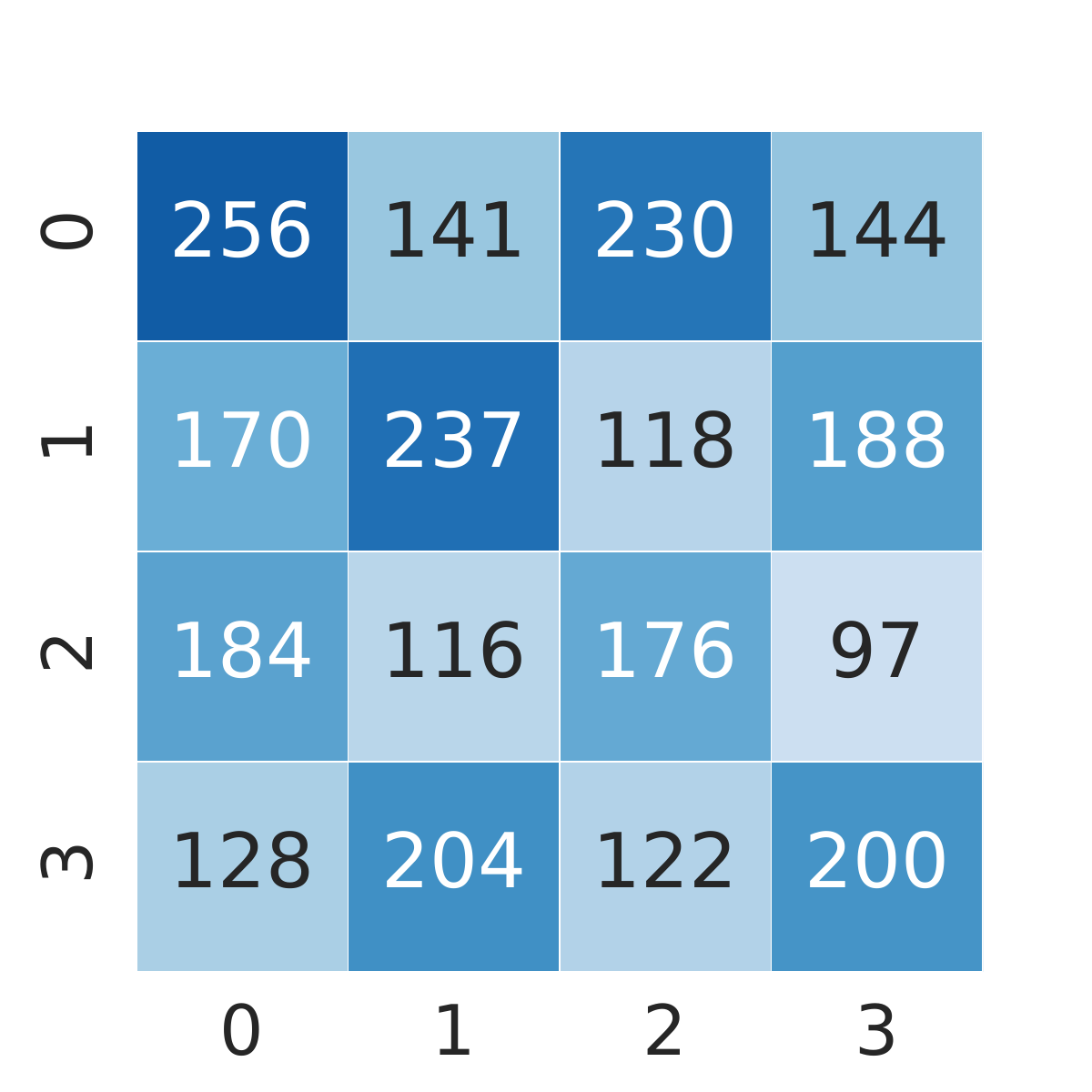}\label{fig:heatmpa-mnist-4x4-100-unif}}
\hspace{0.cm}

\vspace{-0.4cm}

\subfloat[\scriptsize{\Lipi(25)}]
         {\includegraphics[width=0.2\textwidth]{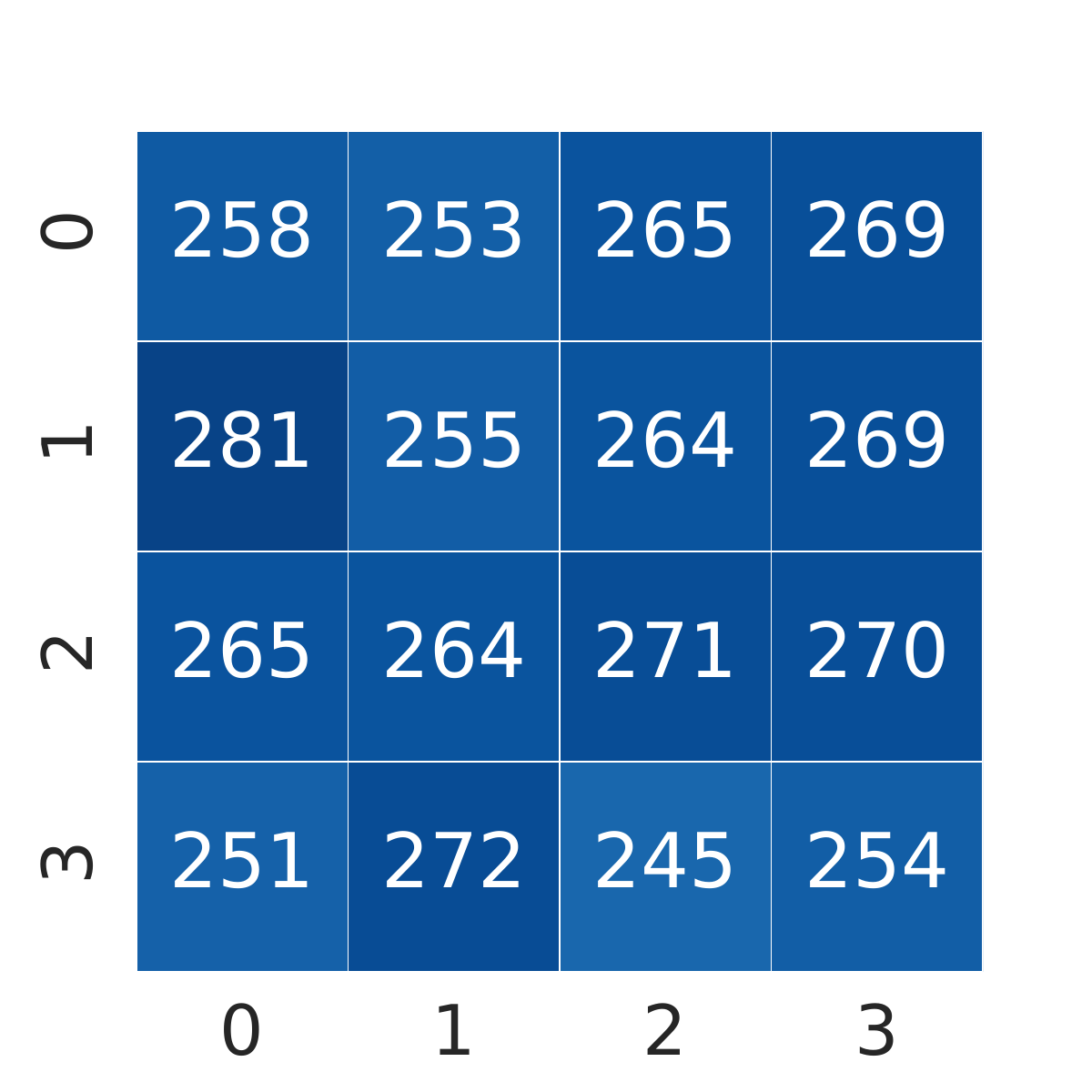}\label{fig:heatmpa-mnist-4x4-025-op}}\hspace{0.1cm}
\subfloat[\scriptsize{\Lipi(50)}]
         {\includegraphics[width=0.2\textwidth]{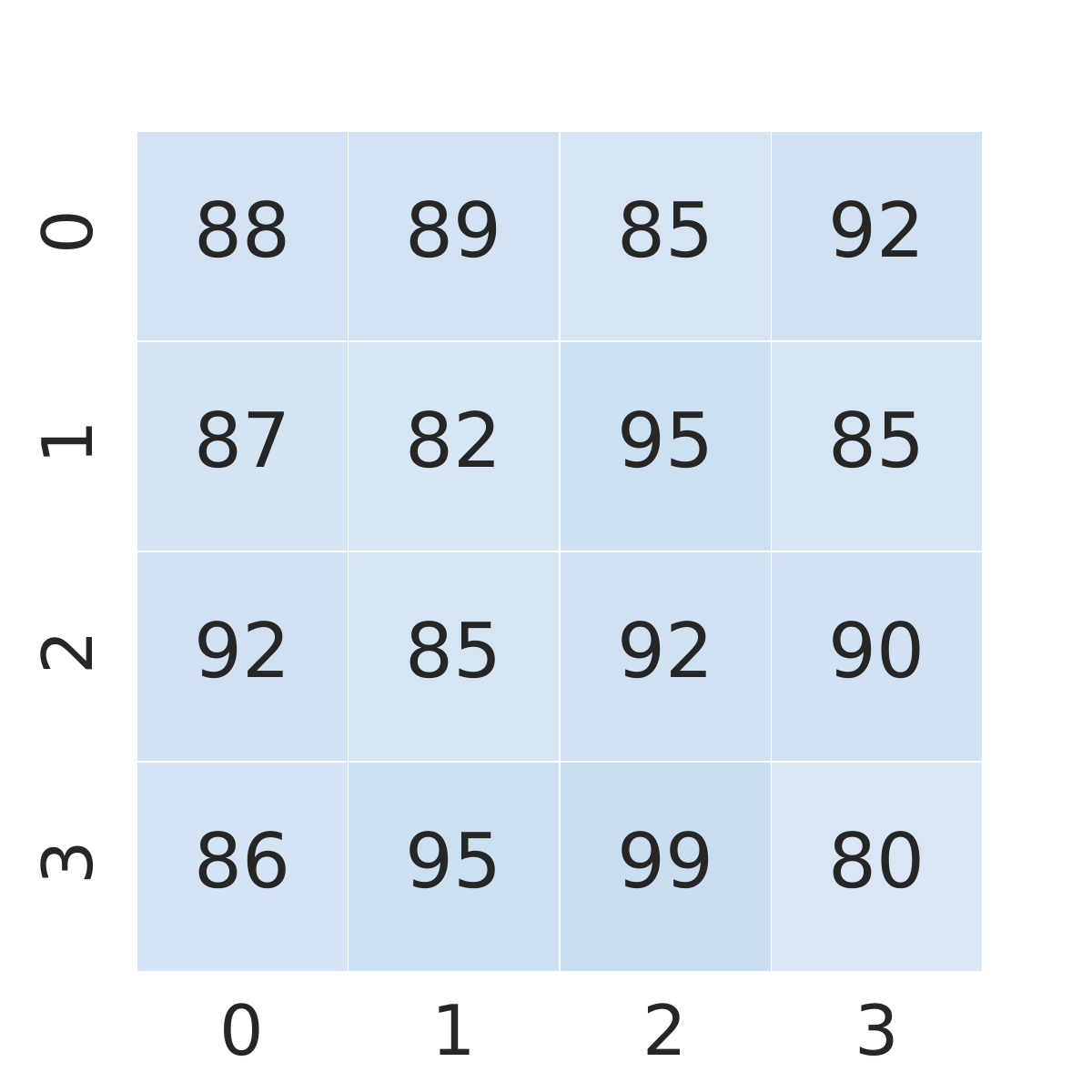}\label{fig:heatmpa-mnist-4x4-050-op}}\hspace{0.1cm}
\subfloat[\scriptsize{\Lipi(75)}]
         {\includegraphics[width=0.2\textwidth]{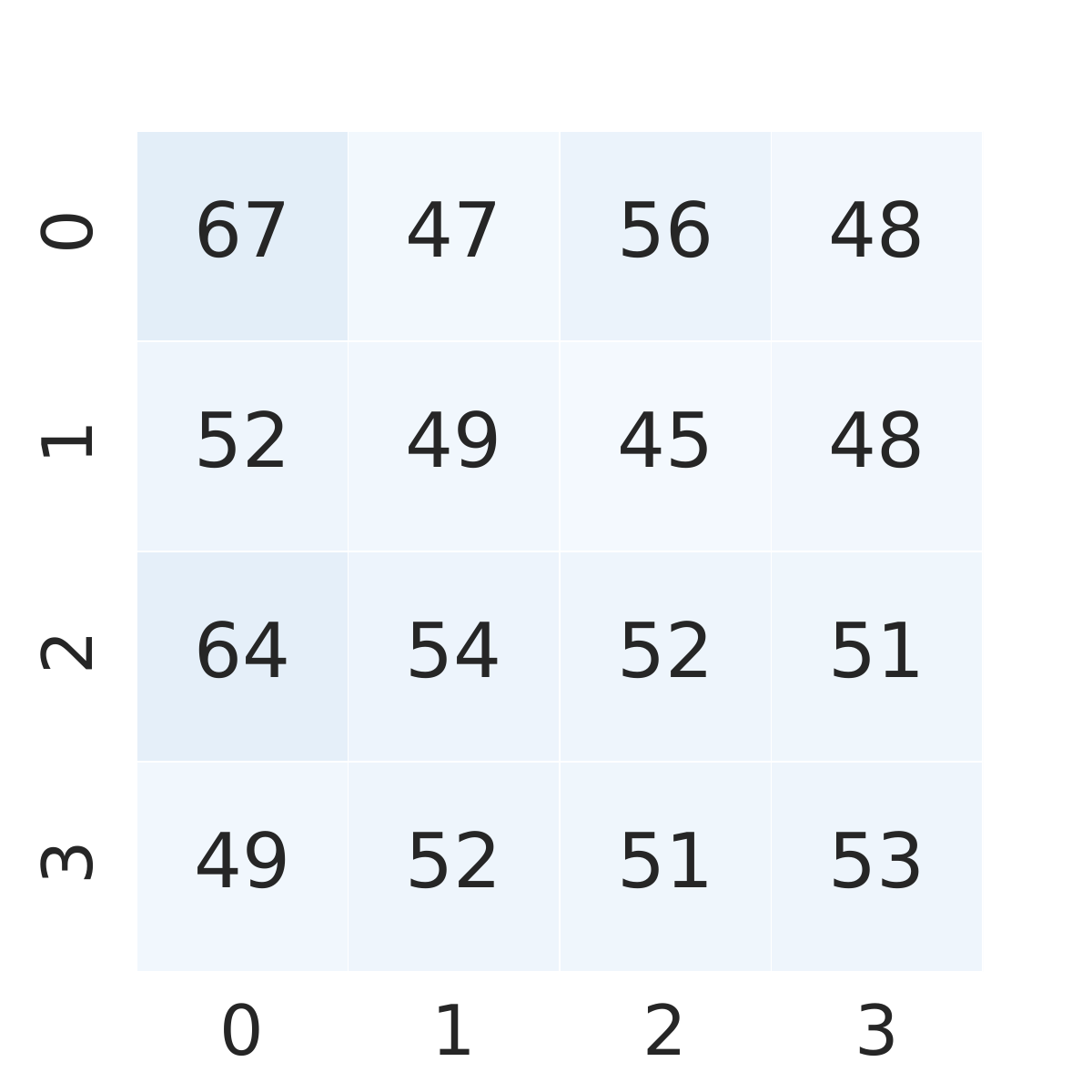}\label{fig:heatmpa-mnist-4x4-075-op}}\hspace{0.1cm}
\subfloat[\scriptsize{\texttt{Lipiz.} (100)}]
         {\includegraphics[width=0.2\textwidth]{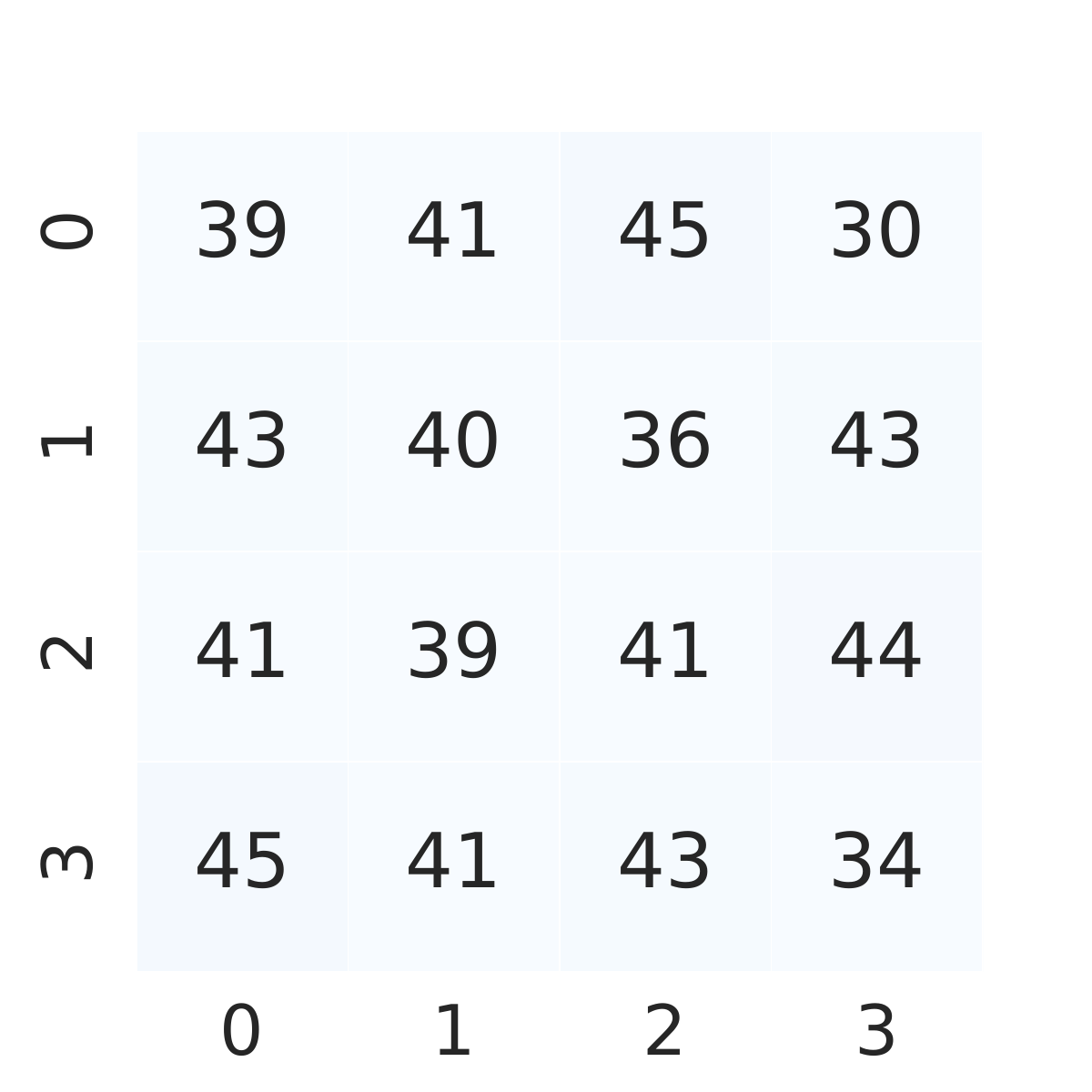}\label{fig:heatmpa-mnist-4x4-100-op}}
  \vspace{-0.2cm}
        \caption{\small{FID score distribution through the grid at different epochs (expresed between parentesis) of an independent run for \spagan and \Lipi in a \xGrid{4} grid. Lighter blues represent lower (better) FID scores. }} 
        \vspace{-0.2cm}
  \label{fig:fids-in-grid}
\end{figure*}

Fig.~\ref{fig:fid-gridsize-evolution} shows the evolution of the
median FID when using \pagan, \spagan, and
\Lipi in a \xGrid{3} grid. 
According to this figure, \Lipi improves the performance of the generators faster than \spagan. 
After the first 75 to 100
training epochs, the FID does not show such a reduction in \Lipi, but
\spagan is able to keep reducing it until the end of the training
process. 
The faster convergence of \Lipi is also illustrated by Fig.~\ref{fig:fids-in-grid} that shows the FID scores in the grid of a given independent run at the epoch number 25, 50, 75, and 100.  This is mainly due to the capacity of exploitation of this method when
using selection/replacement.
\pagan shows a fast FID reduction at the beginning of the
training process. But, the FID sharply oscillates during the first
75 iterations and it converges to worse FID scores than
\Lipi and \spagan.

\noindent

Fig.~\ref{fig:fid-differences} illustrates the differences between
the FID score when using \xGrid{3} and \xGrid{4} or \xGrid{3} and \xGrid{5} for the
same method. This difference for the grid size \xGrid{m} at the $i$
epoch is computed as \hbox{$diff\_FID^{m \times m}_{i}=FID^{3\times
    3}_{i} - FID^{m \times m}_{i}$}.  This figure allows evaluating
the impact on the FID evolution when using different grid (population)
sizes.
%
It shows how \Lipi provides lower FIDs
and it is able to converge faster as the grid size increases during
the first iterations.  In contrast, \spagan gets better FIDs when
increasing the grid size only during the first 50 epochs. \Lipi is
able to take advantage of the diversity generated when the grid size
increase to converge faster to lower FIDs than \spagan.  Thus, in
terms of convergence, selection/replacement helps the convergence of the
coevolutionary training method when exchanging solutions
with the neighborhoods.

\vspace{-0.2cm}
\subsection{Generator Output Diversity} 
\label{sec:tvd}
\vspace{-0.2cm}

The output diversity reports the class distribution of the fake data
produced by a generative model. 
Table~\ref{tab:final-tvds} summarizes the results in terms of TVD.

\begin{table}[!h]
\vspace{-0.4cm}
\setlength{\tabcolsep}{4pt}
\renewcommand{\arraystretch}{0.85}
	\centering
	\small
	\caption{\small TVD results (Low TVD indicates more diversity).}
	\label{tab:final-tvds}
\begin{tabular}{llrrrrr}
\toprule
\textbf{Grid} & \textbf{Method} &  \textbf{Mean}$\pm$\textbf{Std} &  \textbf{Median} &   \textbf{Iqr} &   \textbf{Min} &   \textbf{Max} \\
\midrule
\multirow{4}{*}{\xGrid{3}} & 
\Lipi & 0.12$\pm$0.03 & \textbf{0.12} &  0.04 &  \textbf{0.08}  &  0.18 \\ 
&\spagan & \textbf{0.12$\pm$0.02} & \textbf{0.12} &  0.02 &  \textbf{0.08}  &  \textbf{0.16} \\  
&\nocommgan & 0.83$\pm$0.08 & 0.82 &  0.14 &  0.69  &  0.91 \\ 
&\oppagan & 0.14$\pm$0.02 & 0.14 &  0.04 &  0.10  &  0.19 \\ 
\midrule
\multirow{2}{*}{\xGrid{4}} & 
\Lipi & \textbf{0.11$\pm$0.02} & \textbf{0.11} &  0.02 &  \textbf{0.05}  &  \textbf{0.16} \\ 
&\spagan & 0.12$\pm$0.02 & 0.12 &  0.03 &  0.08  &  \textbf{0.16} \\ 
\midrule
\multirow{2}{*}{\xGrid{5}} &
\Lipi & \textbf{0.10$\pm$0.02} & \textbf{0.11} &  0.02 &  \textbf{0.06}  &  0.16 \\ 
&\spagan  & 0.11$\pm$0.02 & \textbf{0.11} &  0.03 &  0.07  & \textbf{ 0.15} \\ 
\bottomrule
\end{tabular}
\vspace{-0.cm}
\end{table}

For the \xGrid{3} grid experiments, \nocommgan shows the worst (highest) TVD values.
\pagan provides good TVDs but it is statistically less
competitive than \spagan and \Lipi according to the Mann-Whitney U
test.  \spagan and \Lipi show the best results obtaining the same
\textbf{Mean}, \textbf{Median}, and \textbf{Min} values.
Therefore, when using communication between the neighborhoods during the training, the generators are able to produce more diverse data samples.

When increasing the grid size, \spagan and \Lipi improve their results. 
However, \spagan does not
show statistical differences between the same algorithm with different
grid sizes.  
\Lipi \xGrid{5} provides statistically better results than \Lipi  \xGrid{3} and than \spagan. 
The coevolutionary approach used in \Lipi takes advantage of
the divergence generated when increasing the grid size to train
generators that create more diverse data samples.

According to these results and the ones in Section~\ref{sec:final-fid}, we can answer \textbf{RQ1:} 
\textit{What is the effect on the quality of the generators when training with communication or isolation and the presence or absence of selection pressure?}  
Communication and selection pressure allowed \Lipi to converge to generators with the best quality (FID and TVD). 
The diversity resulting from communication resulted in the most competitive results, i.e., \Lipi and \spagan ended with better generators than \nocommgan and \pagan. 
Isolation in training converged to good solutions when the cell was optimizing only one GAN (\pagan). However, when isolation is coupled with a sub-population that applies selection/replacement, the algorithm was not able to converge, and therefore, the quality of the generators returned is the worst.

\vspace{-0.25cm}
\subsection{Genome Space Diversity}
\vspace{-0.2cm}

\begin{figure*}[!h]
\vspace{-0.5cm}
\centering
  \subfloat[\scriptsize{\pagan(\xGrid{3})}]
         {\includegraphics[width=0.2\textwidth]{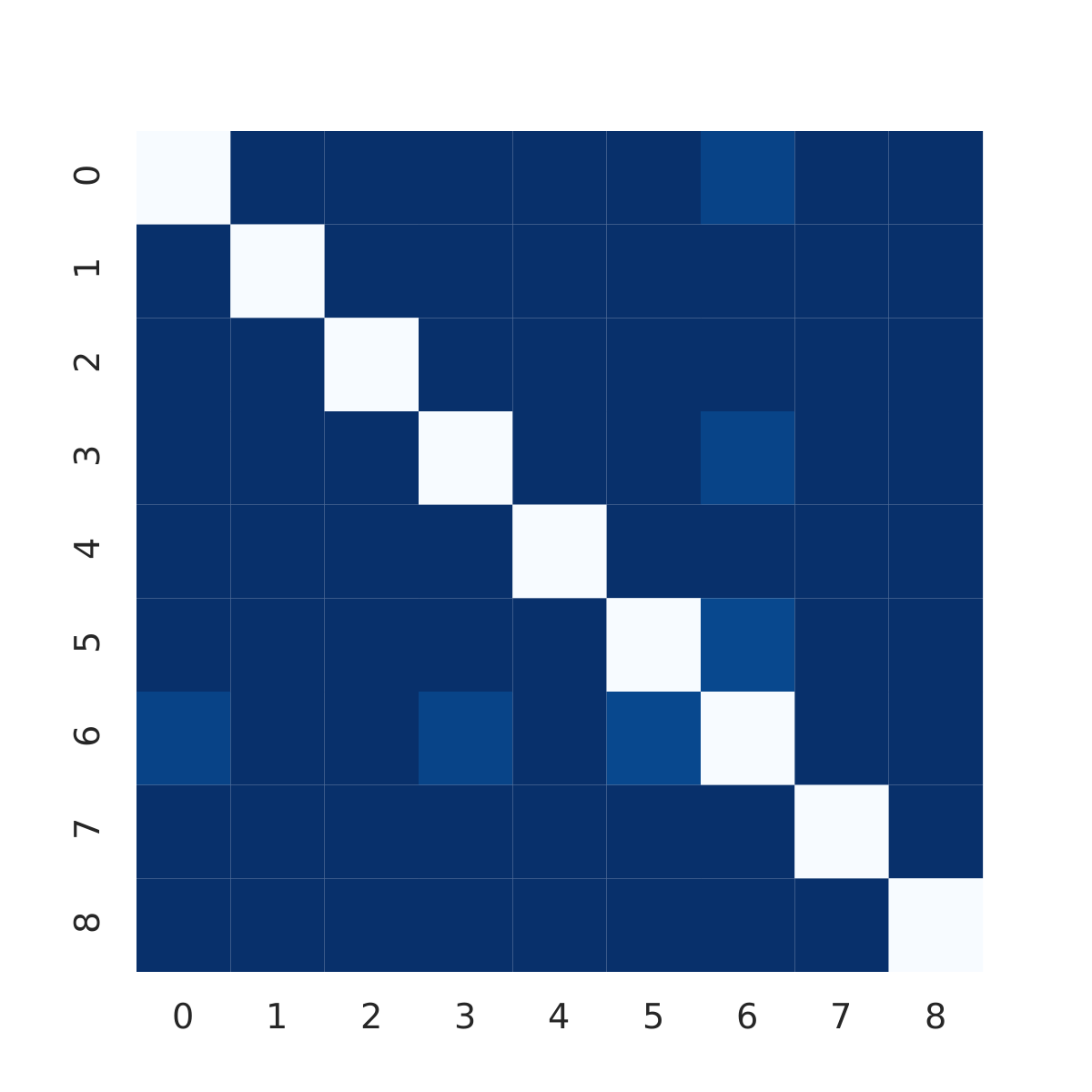}\label{fig:mnist-nets-3x3-pagan}}\hspace{0.1cm}
         \subfloat[\scriptsize{\nocommgan(\xGrid{3})}]
         {\includegraphics[width=0.2\textwidth]{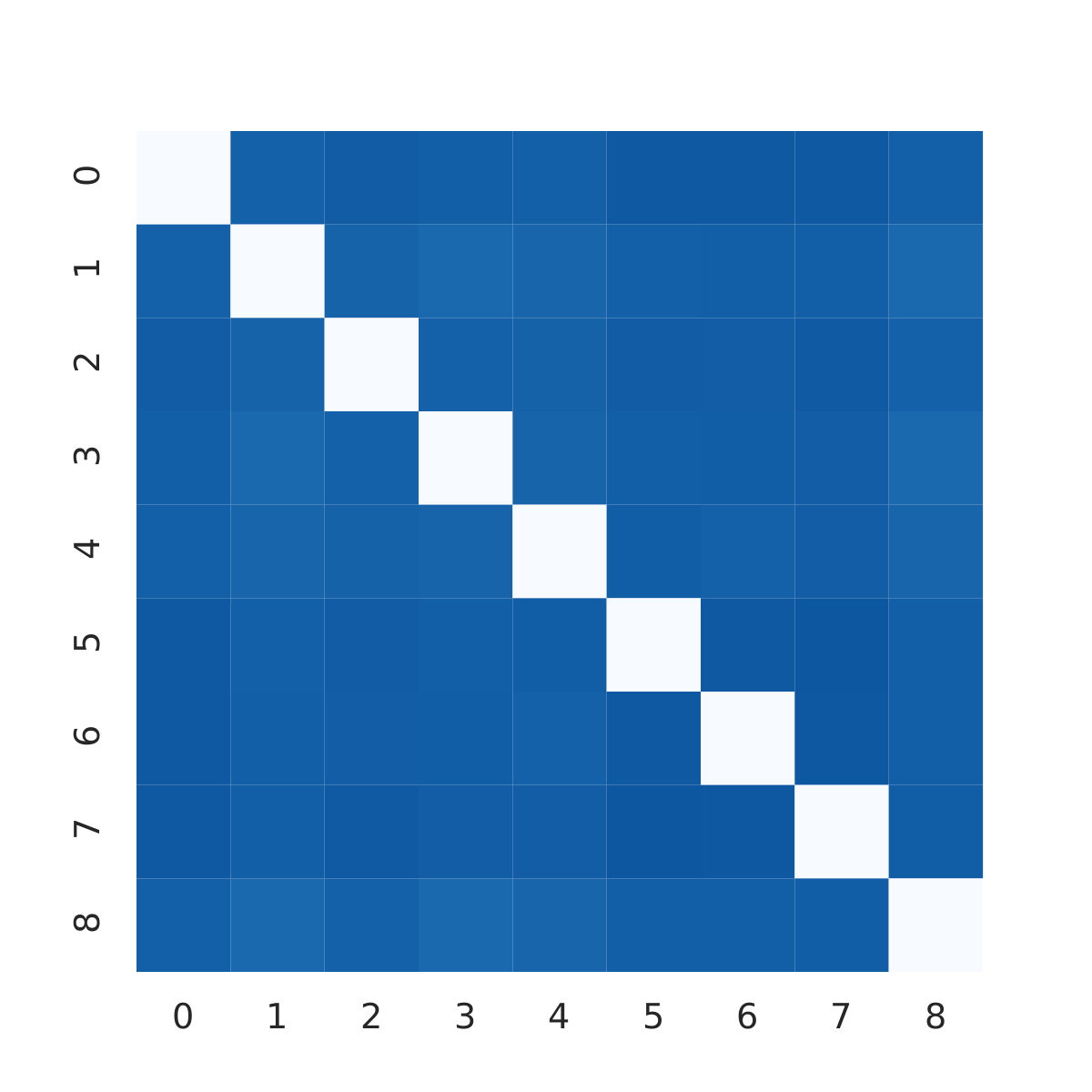}\label{fig:mnist-nets-3x3-nocomgan}} \hspace{0.1cm} 
         \subfloat[\scriptsize{\spagan(\xGrid{3})}]
         {\includegraphics[width=0.2\textwidth]{images/mnist-9-no-selection-d}\label{fig:mnist-nets-3x3-pagan}}\hspace{0.1cm}
    \subfloat[\scriptsize{\texttt{Lipiz.} (\xGrid{3})}]
         {\includegraphics[width=0.2\textwidth]{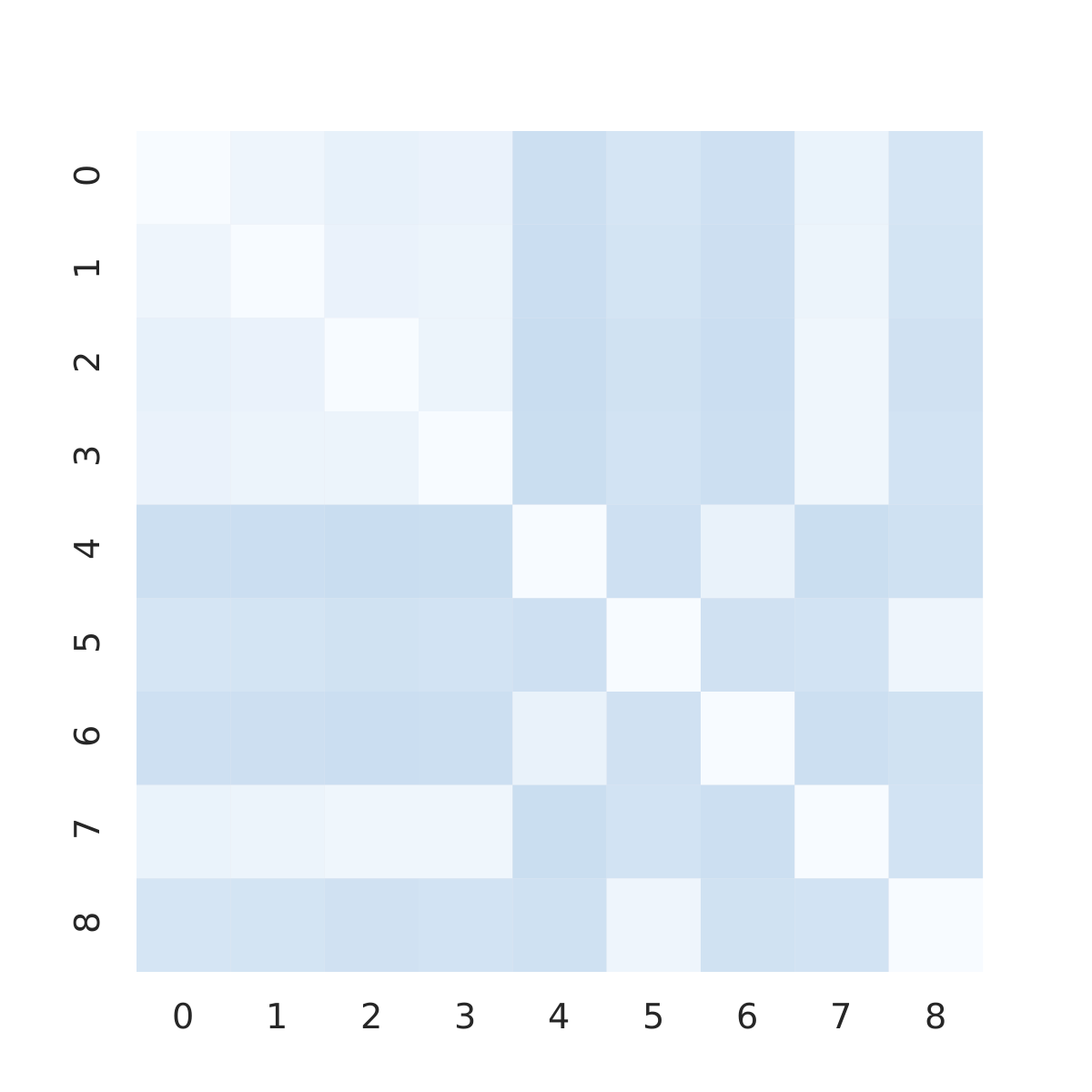}\label{fig:mnist-nets-3x3-lipi}} 
         
\vspace{-0.3cm}        
      
	     \subfloat[\scriptsize{\spagan(\xGrid{4})}]
         {\includegraphics[width=0.19\textwidth]{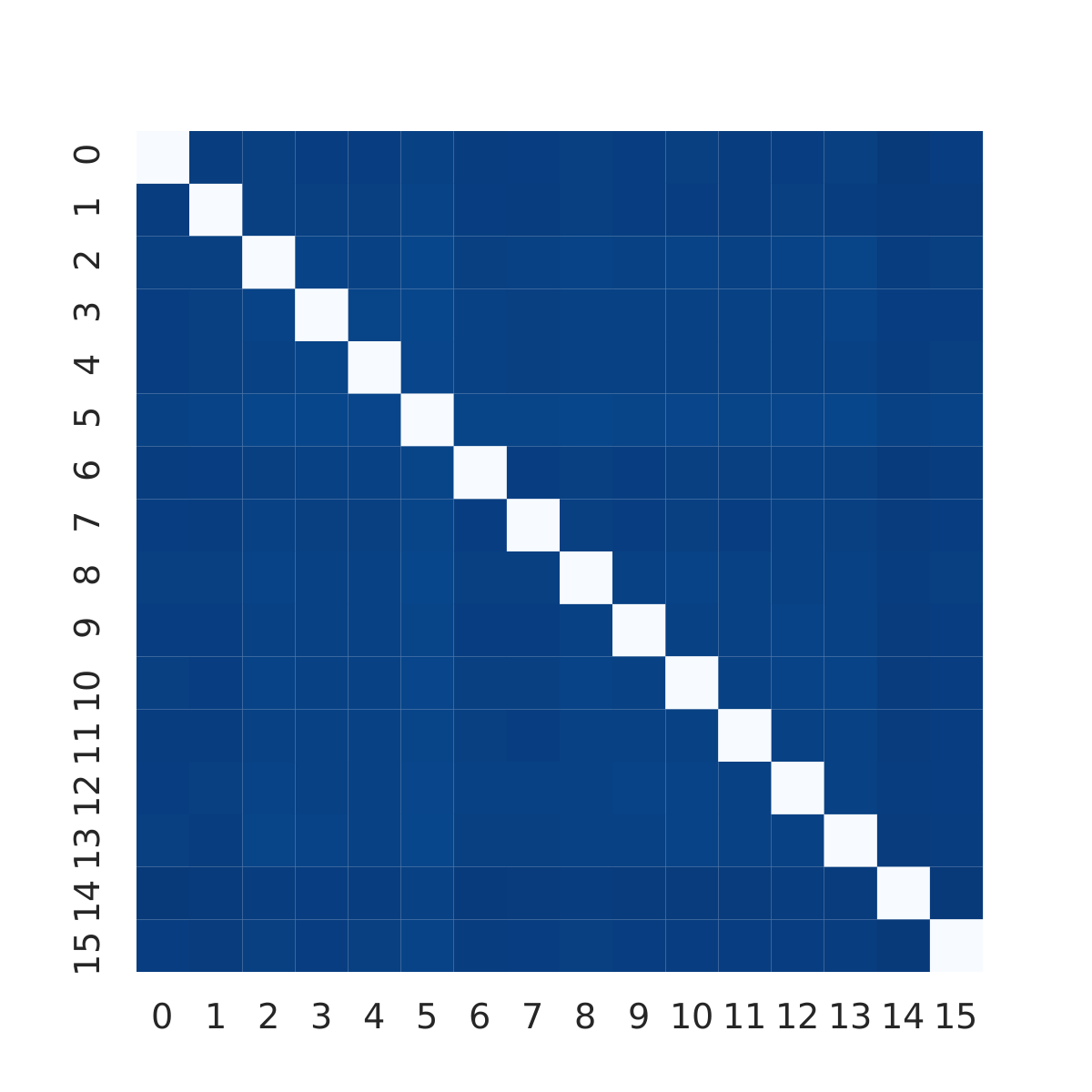}\label{fig:mnist-nets-4x4-pagan}}\hspace{0.1cm}
         \subfloat[\scriptsize{\texttt{Lipiz.} (\xGrid{4})}]
         {\includegraphics[width=0.19\textwidth]{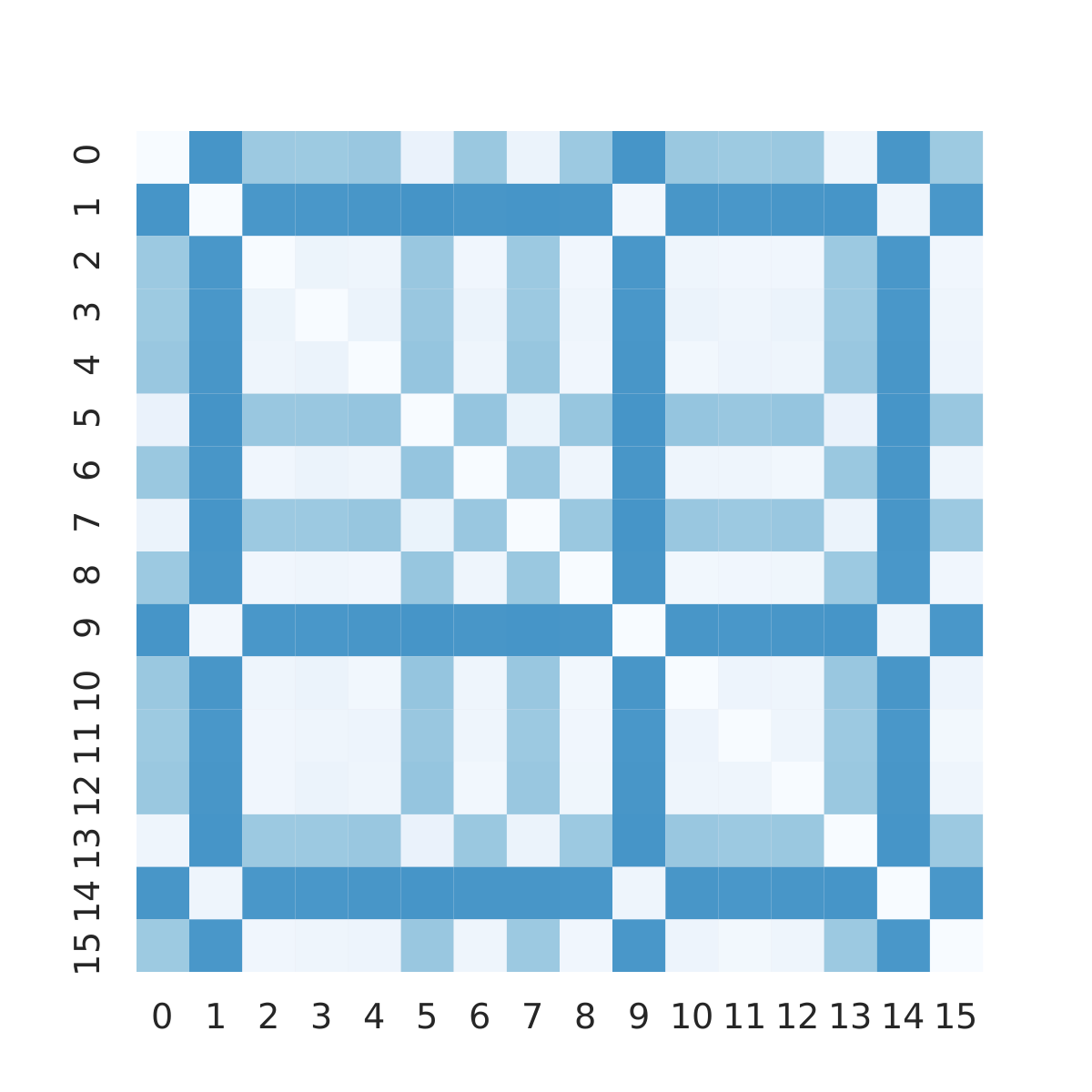}\label{fig:mnist-nets-4x4-lipi}}\hspace{0.1cm}
             \subfloat[\scriptsize{\spagan\xGrid{5}}]
         {\includegraphics[width=0.19\textwidth]{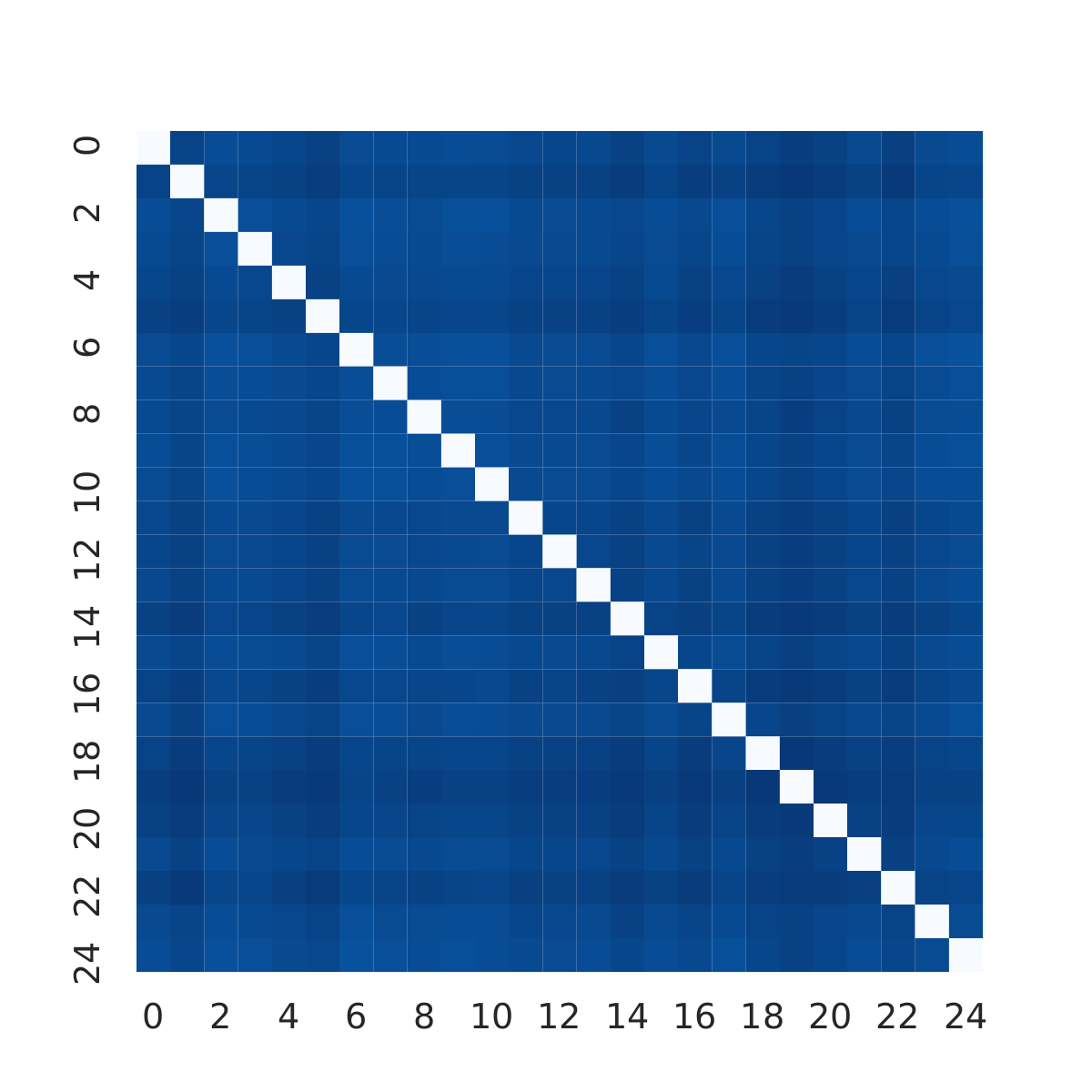}\label{fig:mnist-nets-5x5-pagan}}\hspace{0.1cm}
    \subfloat[\scriptsize{\texttt{Lipiz.} (\xGrid{5})}]
         {\includegraphics[width=0.19\textwidth]{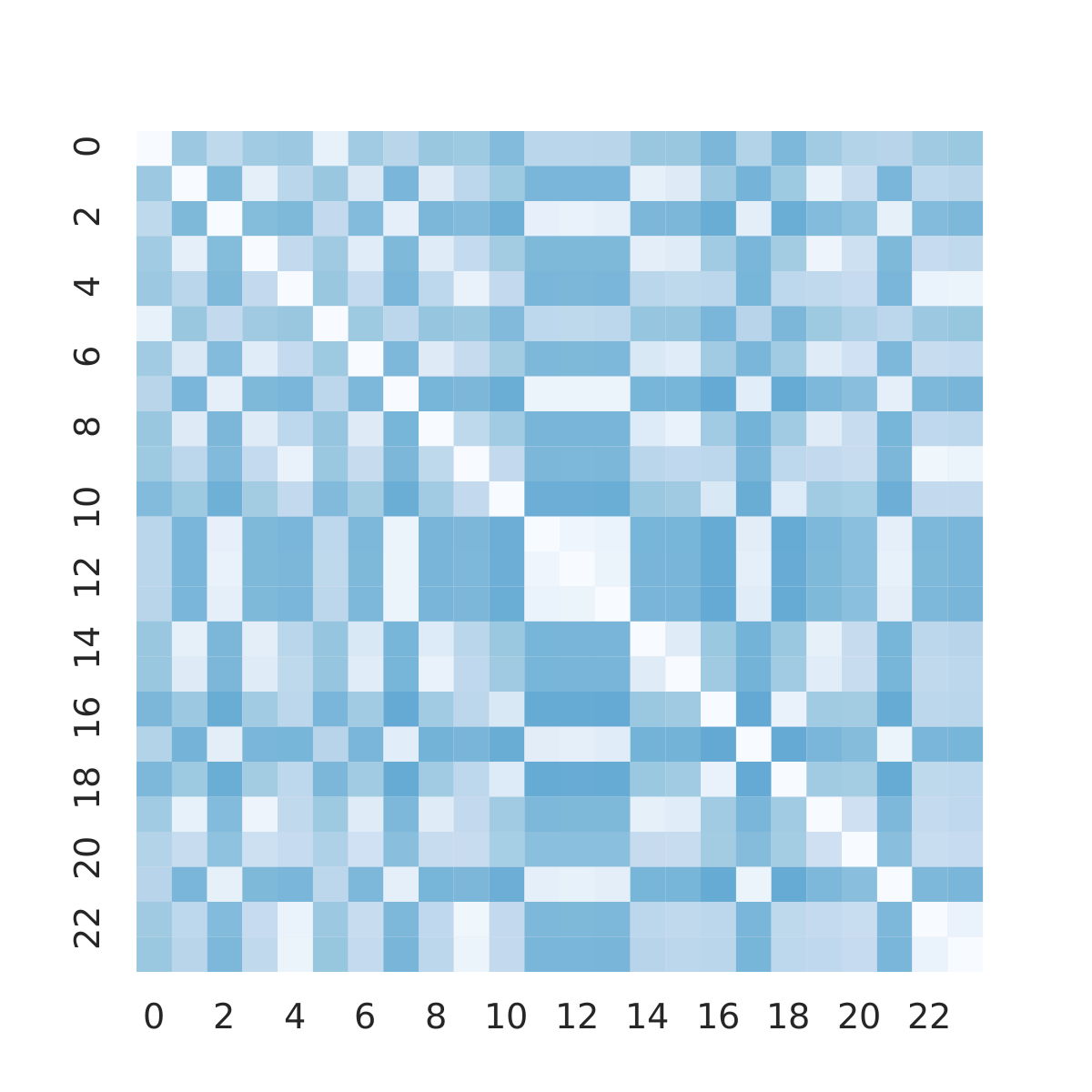}\label{fig:mnist-nets-5x5-lipi}}
           \vspace{-0.2cm}
        \caption{\small{Diversity of population in
            \textit{genome} space. Heatmap of $L_2$ distance between
            the generators at
            the final generation. Dark indicates more diversity. }}
  \label{fig:MNIST_diversity}
\vspace{-0.58cm}
\end{figure*}

We next investigate the diversity of the parameters of the evolved networks. 
Table~\ref{tab:MNIST_diversity} summarizes the $L_2$ distance results for the population at the
end of the independent run that returned the median FID. Fig.~\ref{fig:MNIST_diversity} shows the $L_2$ distances between
all the generators in the grid (x and y axes are the cell number).

\begin{table}[!h]
\vspace{-0.78cm}
\setlength{\tabcolsep}{4pt}
\renewcommand{\arraystretch}{0.85}
	\centering
	\small
	\caption{\small Diversity of population (whole grid) in \textit{genome} space. $L_2$ distances  between the generators at the final generation.}
	\label{tab:MNIST_diversity}
\begin{tabular}{llrrrrr}
\toprule
\textbf{Grid} & \textbf{Method} &  \textbf{Mean}$\pm$\textbf{Std} &  \textbf{Median} &   \textbf{Iqr} &   \textbf{Min} &   \textbf{Max} \\
\midrule
\multirow{4}{*}{\xGrid{3}} 
& \Lipi & 13.85$\pm$8.29 & 18.06 &  15.32 &  4.09  &  23.46 \\ 
& \spagan & 72.70$\pm$25.75 & 81.72 &  2.66 &  78.32  &  84.82 \\
& \nocommgan & 71.07$\pm$25.29 & 79.42 &  5.67 &  74.00  &  86.01 \\ 
& \oppagan & \textbf{109.98$\pm$41.15} & \textbf{128.26} &  27.89 &  \textbf{90.82}  &  \textbf{138.02} \\ 
\midrule
\multirow{2}{*}{\xGrid{4}}& \Lipi & 30.99$\pm$24.28 & 38.09 &  55.82 &  2.72  &  61.34 \\ 
& \spagan & \textbf{87.90$\pm$22.71} & \textbf{93.75} &  1.36 &  \textbf{91.44}  & \textbf{ 95.95} \\ 
\midrule
\multirow{2}{*}{\xGrid{5}} & \Lipi & 31.98$\pm$15.31 & 36.71 &  21.32 &  4.28  &  52.14 \\ 
& \spagan  & \textbf{87.96$\pm$18.06} & \textbf{91.60} &  3.30 & \textbf{ 87.22 } &  \textbf{96.87} \\ 
\bottomrule
\end{tabular}
\vspace{-1.35cm}
\end{table}

The highest generator diversity is shown by \pagan. This is because this method trains a single generator against a single discriminator and it does not use any type of information exchange or communication during the
training process. 
Therefore, each cell converges to different points of the search space. 

\spagan and
\nocommgan provide networks with similar diversity in \xGrid{3}
grid, which is higher than the diversity provided by \Lipi.
This is mainly due to the combination of both communication and the
use of selection/replacement in
\Lipi provokes the sub-populations to converge to similar accurate
individuals, limiting diversity.

The genome diversity in \Lipi is increased when increasing the size of
the grid (see in Fig.~\ref{fig:MNIST_diversity} how the heatmap gets
darker as the grid size is larger).  \spagan increases the $L_2$
distances with the grid size, but in a lower proportion, e.g., from
\xGrid{3} to \xGrid{4} \Lipi increases 123.75\% and \spagan 20.90\%. The diversity in the genome space helps the creation of
ensembles that are able to provide better accuracy. For this reason,
\Lipi can improve results with an increasing grid size (and
diversity).

Answering
\textbf{RQ2:} \textit{What is the effect on the diversity of the network parameters when training with communication or isolation and the presence or absence of selection pressure?} 
The combination of communication and selection pressure permits \Lipi to converge to similar high-quality generators. 
Complete isolation and no-population based GAN training (\pagan) converge to highly different generators, which is expected. 
\spagan illustrates how the absence of selection pressure in the populations keep the individuals diverse in the grid, although there is communication. 
In \xGrid{3} grid experiments, \nocommgan and \spagan have similar diversity but highly different quality. This shows how maintaining diversity is not enough to ensure robust GAN training.

\vspace{-0.2cm}
\subsection{Computational Efficiency}
\vspace{-0.1cm}

Now, we analyze the computational efficiency of the GAN training
methods, taking into account that they use the same number of
training epochs but used different components. 
All these methods apply asynchronous parallelism and the time required by each cell of the grid to perform the same number of training epochs varies.
Thus, we report the computational time of a run as the time required
by each independent run to finish and return the best ensemble of generators found by all the cells. 

\begin{table}[!h]
\vspace{-.6cm}
\setlength{\tabcolsep}{4pt}
\renewcommand{\arraystretch}{0.85}
	\centering
	\small
	\caption{\small Computation time in minutes.}
	\label{tab:computation-time}
\begin{tabular}{llrrrrr}
\toprule
\textbf{Grid} & 
\textbf{Method} &  \textbf{Mean}$\pm$\textbf{Std} &  \textbf{Median} &   \textbf{Iqr} &   \textbf{Min} &   \textbf{Max} \\
\midrule
\multirow{4}{*}{\xGrid{3}} & 
\Lipi & 87.89$\pm$1.15 & 87.73 &  1.23 &  85.57  &  90.23 \\
&\spagan & 87.20$\pm$0.31 & 87.15 &  0.34 &  86.72  &  88.02 \\
&\nocommgan & 81.88$\pm$4.55 & 81.34 &  9.41 &  74.40  &  88.19 \\ 
&\oppagan & \textbf{38.07$\pm$2.73} & \textbf{37.51} &  3.49 &  \textbf{33.57}  &  \textbf{44.28} \\ 
\midrule
\multirow{2}{*}{\xGrid{4}} & 
\Lipi & 91.30$\pm$0.94 & 91.07 &  1.01 &  90.23  &  94.26 \\
& \spagan & \textbf{90.72$\pm$0.58} & \textbf{90.89} &  0.39 &  \textbf{88.97}  & \textbf{ 91.27} \\

\midrule
\multirow{2}{*}{\xGrid{5}} &
\Lipi & 105.64$\pm$3.25 & 107.22 &  4.48 &  100.25  &  111.05 \\ 
& \spagan & \textbf{101.88$\pm$1.64} & \textbf{100.91 }&  2.10 &  \textbf{100.18}  &  \textbf{105.52 }\\  
\bottomrule
\end{tabular}
\vspace{-0.3cm}
\end{table}

The computational time of \oppagan includes the whole process, i.e., the time of training the GANs plus the time of optimizing the ensemble weights by using the ES-(1+1). 
This method requires the shortest run time (see Table~\ref{tab:computation-time}). 
This is because \oppagan trains a single network in each cell and there is no communication among the different cells.
According to the Mann-Whitney U and posthoc statistical tests,
\Lipi requires the longest computation times.
It employs both communication and selection/replacement algorithm components.

When comparing among \Lipi, \spagan, and \nocommgan,
the impact on the computation time of applying communication in \Lipi and \spagan is higher than the use of selection/replacement in \Lipi and \nocommgan (see Table~\ref{tab:computation-time} \textit{\xGrid{3}-Grid}). 
This increase in the computation cost may increase in problems that require the use and exchange among cells of bigger models (networks) for the generators and discriminators.

Therefore, answering \textbf{RQ3:} \textit{What is the impact on the computational cost of applying migration and selection/replacement?}, the effect of applying selection/replacement is negligible when comparing it with the impact of communicating among the cells, when running the methods on \xGrid{3} grid. However, it increases as the grid size is larger. 
Moreover, when bigger networks need to be trained to handle harder problems, they will have many more parameters and more need for communications.

\vspace{-0.4cm}
\section{Conclusions and Future Work}
\label{sec:conclusions}
\vspace{-0.1cm}
We have empirically shown that the spatially distributed coevolutionary training applied by \Lipi is the best choice 
among options with/without communication and selection/replacement components to train GANs. 
The combination of selection pressure that promotes convergence in the sub-populations and communication with the overlapped neighborhoods maintains enough diversity to robustly train the networks in the sub-population. 
Moreover, the use of these two operations does not entail a very significant increase in the computation time (about four minutes in the larger grid size).

\spagan illustrates the importance of the communication among the cells (i.e., fostering diversity). It is able to converge to high-quality solutions (generators) although it does not apply selection. This emphasizes the value of exchanging the best individuals even they are just used to train the center of the cells.  \nocommgan provides the least competitive results. 
Training the networks with a coevolutionary flavor does not ensure convergence, even when it is done in sub-populations and it uses selection/replacement.  

Future work will include further analysis of coevolutionary GAN training in other benchmarks (i.e., problems and data sets). 
We will evaluate the scalability of this type of training by using larger grids.  
We will study the effect of training the GANs with different kinds of loss functions. 
We will assess the impact on the robustness when using different types of neighborhoods. 
Finally, we will analyze the evolution of the network weights through the generations to better understand the dynamics of this type of GAN training.

\section*{Acknowledgments}
This research was partially funded by European Union’s Horizon 2020
research and innovation program under the Marie Skłodowska-Curie grant agreement No 799078, by the Junta de Andaluía UMA18-FEDERJA-003, European Union H2020-ICT-2019-3, and the Systems that Learn Initiative at MIT CSAIL.

\bibliographystyle{splncs04}
\bibliography{bibliography} 
\end{document}